\pdfoutput=1
\documentclass[10pt,twocolumn,letterpaper]{article}

\usepackage[pagenumbers]{cvpr}

\usepackage{microtype}

\makeatletter
\renewcommand\paragraph{\@startsection{paragraph}{4}{\z@}{.2em \@plus.2em \@minus.1em}{-.8em}{\normalfont\normalsize\bfseries}}
\g@addto@macro\normalsize{%
  \setlength\abovedisplayskip{5pt \@plus2pt \@minus2pt}%
  \setlength\belowdisplayskip{5pt \@plus2pt \@minus2pt}%
  \setlength\abovedisplayshortskip{0pt \@plus2pt}%
  \setlength\belowdisplayshortskip{3pt \@plus2pt \@minus1pt}}
\makeatother
\usepackage{colortbl}

\usepackage{booktabs}
\usepackage{tabularx}
\usepackage{algorithm}
\usepackage{algpseudocode}

\definecolor{cvprblue}{rgb}{0.21,0.49,0.74}
\usepackage[pagebackref,breaklinks,colorlinks,allcolors=cvprblue]{hyperref}

\title{What Places an Action Boundary? Structure-Aware Attribution for Temporal Action Localization}

\author{
Ruoshan Lan$^{1}$ \qquad
Zongru Shao$^{2}$ \qquad
Chieh Tsai$^{3}$ \qquad
Mihai Surdeanu$^{1}$\\[6pt]
$^{1}$Department of Computer Science, University of Arizona\\
$^{2}$Silicon Austria Labs\\
$^{3}$Department of Electrical and Computer Engineering, University of Arizona\\[4pt]
{\small\fontencoding{T1}\fontfamily{lmtt}\selectfont \{ruoshanlan, vegetableclean, msurdeanu\}@arizona.edu}\\
{\small\fontencoding{T1}\fontfamily{lmtt}\selectfont zongru.shao@silicon-austria.com}
}

\begin{document}
\maketitle
\begin{abstract}
Temporal action localization (TAL) detectors must identify actions and estimate their temporal boundaries, but localization remains a major source of detection errors. 
While understanding what determines boundary placement requires faithful attribution of temporal evidence, existing attribution methods have not sufficiently addressed the temporal structure of TAL or the need to tailor attribution to boundary localization.
We introduce a structure-aware attribution framework that averages contributions across random evidence-restoration orders following the Shapley--Shubik principle. 
We make it tractable by using the detector's receptive field to limit which time steps are ordered individually, and by approximating each order with a soft gradient path.
For boundary predictions, we further introduce cell-structured restoration orders and boundary-anchored references that selectively remove adjacent evidence.
We also adapt faithfulness evaluation to assess the joint effects of contiguous temporal evidence and the direction of boundary displacement. Experiments on THUMOS14 and EPIC-KITCHENS-100 with TriDet and ActionFormer demonstrate improved attribution faithfulness over existing methods. 
Attribution analysis and controlled interventions further reveal that class-score changes do not reliably track boundary displacement, providing input-level evidence that classification relies on content less informative about boundary placement.
Code and evaluation artifacts will be publicly released.
\end{abstract}    
\section{Introduction}
\label{sec:intro}

Temporal action localization (TAL) is a fundamental task in video understanding, with many downstream applications~\cite{vahdani2022deep}. 
For each action instance in an untrimmed video, a TAL detector predicts its class label together with its start and end times.
Of these outputs, the start and end times are the bottleneck: localization error was found to be the most significant error, and predictions with localization errors can still be among the highest-scored~\cite{alwassel2018diagnosing, zeng2024benchmarking}. 
Detector designs attribute this to the classifier relying on discriminative content that is less informative about boundary placement~\cite{zhao2017ssn,liu2022tadtr,hu2022estimation}.
However, to our knowledge, this explanation has not been examined at the level of input evidence. 
Even with continued advances in TAL detector design, changes to temporal aggregation often yield only modest gains under controlled comparisons~\cite{liu2025opentad}. 
These findings motivate a closer examination of how temporal evidence contributes to classification and, in particular, boundary placement, calling for attribution methods that faithfully explain both outputs while respecting the temporal structure of video inputs and TAL detectors.

Simplifying a bit, TAL can be seen as the temporal counterpart of object detection. In object detection, attribution methods explain individual predictions through their class scores and bounding-box coordinates~\cite{zhao2023odam,nakai2025igodam}. We study a similar attribution task in TAL, attributing an action instance's class score and temporal boundaries to the input snippet embeddings. 

Explaining TAL detections faces one challenge and one gap. 
The challenge is that a snippet's contribution depends on its temporal context.
First, neighboring snippets are temporally redundant: they carry overlapping evidence, so masking one can leave its information available in the others~\cite{leung2023temporal, jang2025timing}. 
Second, TAL detectors draw on temporal context beyond the action itself, from the surroundings of a proposal and its  boundaries~\cite{zhao2017ssn,chen2022dcan,liu2024adatad} to background and video-level semantics~\cite{zeng2019graph,xu2020gtad}, and this context improves their mean average precision (mAP). 
Because of these dependencies, a snippet may contribute little when redundant evidence is present but become important when that evidence is absent. Straight-line integrated gradients (IG)~\cite{sundararajan2017axiomatic} restores all snippets simultaneously along a single path, limiting its ability to capture such context-dependent contributions.

The mentioned gap concerns boundary attribution. 
Although existing detector attribution methods explain both classification and localization outputs, they largely apply the same attribution design to both~\cite{kuroki2024bsed,mi2026eelris,nakai2025igodam,nakai2026adaptive,yamauchi2026vxcode}. 
However, boundary localization is a distinct task. 
Unlike a class score, a boundary position is a continuous output that can shift in either temporal direction and whose explanation depends on the reference it is compared against~\cite{letzgus2022toward}, calling for attribution designs tailored to boundary placement. 
Faithfulness evaluation faces a related gap. Existing detection attribution tests evaluate localization using unsigned measures, such as absolute boundary displacement~\cite{mi2026eelris}, which cannot distinguish evidence that moves a boundary earlier from evidence that moves it later. Although direction-aware tests exist for regression attribution, they have not been adapted to detection task for boundary's faithfulness evaluation.

To address them, we introduce a structure-aware attribution framework for TAL. 
Instead of attributing temporal evidence along a single path, we average contributions over different restoration orders to capture their dependence on temporal context, while exploiting the detector's structure to keep computation tractable. 
We further tailor attribution to boundary localization, treating boundary placement as a distinct explanation target from classification. 
We evaluate the resulting attributions with faithfulness tests adapted to temporal evidence and directional boundary changes, and use them to investigate how detectors place action boundaries.
Our main contributions are:
\begin{itemize}
\item \textbf{Efficient random-order path attribution for TAL.}
We develop two computational approximations that make random-order attribution tractable: a receptive-field window that maintains full-input coverage through a context unit, and soft restoration paths whose evaluation cost per order is independent of the number of ordered units.
\item \textbf{Boundary-specific path and baseline.}
We tailor attribution to boundary predictions through cell-structured restoration orders and boundary-anchored references that selectively remove adjacent evidence.
\item \textbf{Faithfulness evaluation and validation.}
We adapt regression-based insertion and deletion tests to directional boundary displacement and extend faithfulness correlations to contiguous temporal blocks. Using these metrics, we validate our methods on THUMOS14 and EPIC-KITCHENS-100 with TriDet and ActionFormer.
\item \textbf{Attribution-based analysis of boundary placement.}
Combining attribution analysis with controlled interventions, we show that class-score changes do not reliably track boundary displacement, providing input-level evidence that classification relies on content less informative about boundary placement.
\end{itemize}

\section{Related Work}
\label{sec:related_work}

\paragraph{Detection attribution and temporal action localization.}
Object-detection explainers attribute individual predictions through class scores or bounding-box coordinates. White-box methods use gradients or integrated gradients~\cite{zhao2023odam,nakai2025igodam,nakai2026adaptive}, sometimes averaged over noisy inputs or multiple references~\cite{padmanabhan2023dext,kien2026enhancing}, while black-box methods use random masks~\cite{petsiuk2021drise} or cooperative-game formulations~\cite{kuroki2024bsed,yamauchi2026vxcode}. EEL-RIS~\cite{mi2026eelris} further treats individual box sides as explanation targets and introduces localization-specific faithfulness tests; like those of D-RISE~\cite{petsiuk2021drise}, these tests measure unsigned changes such as intersection over union (IoU) or absolute displacement and thus cannot tell whether evidence moves a boundary earlier or later.

In TAL, by contrast, interpretation has mainly served other purposes: the saliency of a video classifier has been used to localize actions~\cite{bargal2018excitation}, Action Sensitivity Learning~\cite{shao2023action} learns how much each frame contributes to classification and to localization and uses these weights to recalibrate training, and Ahn \etal~\cite{ahn2024interpretable} visualize a detector's own per-frame class probabilities to interpret action transitions.
None of these works attributes the class score or the start and end times of an individual detection to the input snippets. We do so and test whether the attributions are faithful to the detector.

\paragraph{Time-series attribution.}
Time-series explainers address temporal dependencies through learned masks and replacement strategies~\cite{crabbe2021explaining,enguehard2023learning,liu2024explaining}, or Shapley-based grouping of steps into windows or segments to reduce complexity or preserve joint evidence~\cite{bento2021timeshap,nayebi2023windowshap,kim2026groupsegment}. TIMING~\cite{jang2025timing} uses segment-based masking with integrated gradients, while faithfulness evaluations also perturb contiguous subsequences~\cite{schlegel2019towards,simic2025comprehensive}. These methods explain sequence-, window-, or time-step-level predictions rather than detected segments and their boundaries. Our grouping instead follows the detector's feature pyramid, the feature sequences at several temporal resolutions on which the detector predicts actions of different lengths~\cite{zhang2022actionformer,shi2023tridet}.

\paragraph{Path and baseline design.}
Integrated Gradients (IG)~\cite{sundararajan2017axiomatic} integrates gradients along a straight-line path from a reference to the input, corresponding to the Aumann--Shapley allocation. Subsequent work explores both path and baseline design.
For path design, Guided IG~\cite{kapishnikov2021guided} adapts the path using gradients, SPI~\cite{jeon2023beyond} averages over randomized monotone paths, and Spectral IG~\cite{kim2026spectral} restores ordered spectral components with controllable overlap. A related approach averages marginal contributions across restoration orders, yielding the Shapley--Shubik allocation or Baseline Shapley when missing features are replaced by reference values~\cite{sundararajan2017axiomatic,sundararajan2020many}. This averaging is typically approximated by sampling permutations~\cite{mitchell2022sampling}. Feature grouping further constrains restoration orders, as in the Owen value~\cite{owen1977values,casajus2008characterization} and hierarchical explainers~\cite{teneggi2023fast,bento2021timeshap}.
For baseline design, Expected Gradients~\cite{erion2021improving} averages over data references, Shapley Integrated Gradients~\cite{liu2023new} constructs references from Shapley coalitions, and IG$^2$~\cite{zhuo2024ig2} jointly designs counterfactual baselines and paths.

Our method also averages over restoration orders, but groups steps by the detector's structure and adapts both path and baseline design to boundary outputs, to better tailor these designs to the detection task and its localization sub-task.

\section{Method}
\label{sec:method}

\begin{figure*}[t]
    \centering
    \includegraphics[width=\textwidth]{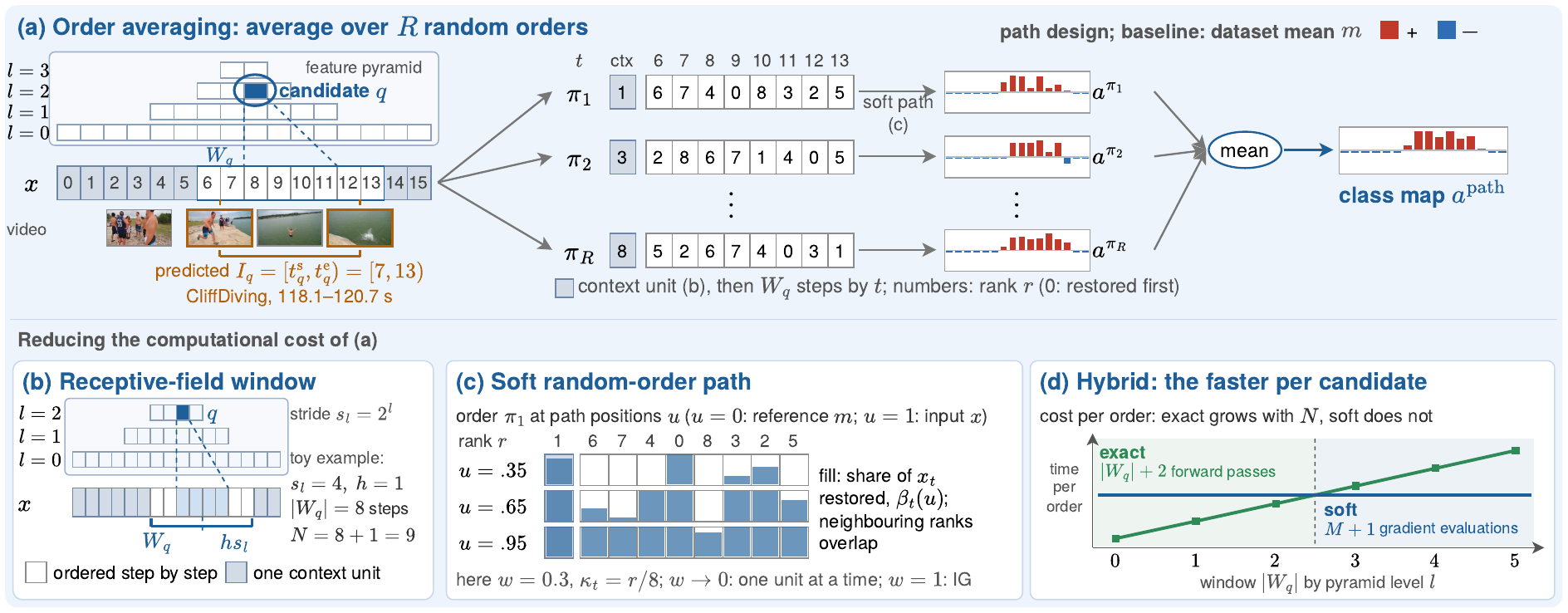}
        \caption{
    \textbf{Random-order path attribution}, shared by all targets and used
    directly as the class map (\cref{sec:path}).
    (a) A candidate $q$ is one position of the detector's feature pyramid;
    the example is a TriDet detection on THUMOS14 (CliffDiving,
    118.1--120.7\,s; start and end frames outlined). Its path attributions
    $a^{\pi}$ are averaged over $R$ random orders of the units into
    $a^{\mathrm{path}}$; the baseline is the dataset mean $m$.
    Two approximations reduce the cost:
    (b) only the steps in the receptive-field window $W_q$ are ordered independently,
    and (c) each order is integrated along a soft path whose cost does not
    depend on the number of units $N$.
    (d) Each candidate uses the faster of exact forward differences and the
    soft path, which for the wide windows of coarse pyramid levels is the soft
    path.
    }

    \label{fig:path-units}
\end{figure*}

\begin{figure*}[t]
    \centering
    \includegraphics[width=\textwidth]{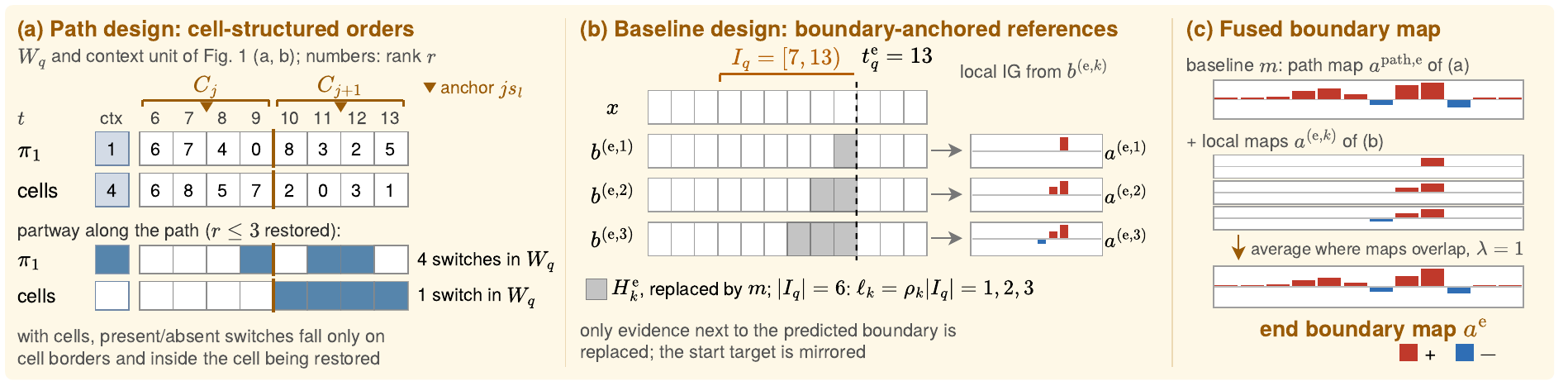}
       \caption{
    \textbf{Boundary-specific path and baseline} (\cref{sec:boundary-map}),
    on the toy axis of \cref{fig:path-units}.
    (a) Path design: for start and end, the orders keep the steps of each
    pyramid cell together, so present/absent switches in partly restored
    inputs fall mainly on cell borders.
    (b) Baseline design: the references $b^{(\mathrm{e},k)}$ replace the nested
    spans $H_k^{\mathrm{e}}$ just inside the predicted end with the mean $m$
    (end target; the start is mirrored).
    (c) Their local maps $a^{(\mathrm{e},k)}$ are averaged with the
    cell-structured path map $a^{\mathrm{path},\mathrm{e}}$ where they overlap
    (\cref{eq:combine-map}, $\lambda=1$).
    }

    \label{fig:base-trunc}
\end{figure*}

\subsection{Problem Formulation}
\label{sec:prelim}

We explain a frozen dense TAL detector operating on pre-extracted snippet features $x=(x_0,\ldots,x_{T-1})\in\mathbb{R}^{T\times D}$, where $x_t$ represents one input step. The detector constructs a temporal feature pyramid with stride $s_l=2^l$ at level $l$. At each pyramid position $p$, a classification head predicts action scores and a localization head predicts boundaries.

We explain a candidate $q=(l,p,c)$, taken before non-maximum suppression (NMS) and indexed by pyramid level $l$, position $p$, and action class $c$. Its predictions define three scalar attribution targets: the class score $F_q^{\mathrm{cls}}\in(0,1)$ and the decoded start and end times $F_q^{\mathrm{s}},F_q^{\mathrm{e}}$ in seconds (\cref{fig:path-units}a).
For example, the candidate $q$ in \cref{fig:path-units}a is a level-2 TriDet prediction of CliffDiving from 118.1 to 120.7\,s, drawn as the interval $[7,13)$ on a toy axis of 16 input steps.

Following~\cite{padmanabhan2023dext,jiang2023nagigos}, we evaluate these same candidate outputs on every perturbed or interpolated input. This keeps each attribution target a fixed function of the input, whereas re-matching detections, i.e., rerunning detection and selecting the most similar detection on each perturbed input~\cite{nakai2025igodam,petsiuk2021drise,kuroki2024bsed}, may switch the explained prediction.

We denote the corresponding clean (unperturbed) predicted interval on the input-step axis by $I_q=[t_q^{\mathrm{s}},t_q^{\mathrm{e}})$ and use the dataset-mean snippet feature $m$ to replace removed input steps. Boundary attributions are signed, with positive and negative values indicating contributions toward later and earlier boundary positions, respectively.

\subsection{Random-Order Path Attribution for TAL}
\label{sec:path}

\subsubsection{Averaging over Temporal Contexts}
\label{sec:path-avg}

To account for the dependence of temporal evidence on its context (\cref{sec:intro}), we average contributions over random orders of evidence restoration rather than relying on a single interpolation path. 
This follows the Shapley--Shubik principle, which treats input units as players and averages their marginal contributions across permutations. 
Let $v(S)$ denote the target when units in $S$ retain their input values and all others take the baseline $m$, as in Baseline Shapley~\cite{sundararajan2020many}. The attribution of unit $i$ is
\begin{equation}
\phi_i
=
\mathbb{E}_{\pi}\bigl[v(S_\pi^i\cup\{i\})-v(S_\pi^i)\bigr],
\label{eq:shapley-order}
\end{equation}
where $S_\pi^i$ contains the units preceding $i$ in a uniformly random order $\pi$.
For example, $\pi_1$ in \cref{fig:path-units}a restores step 9 first, so it measures the contribution of step 9 alone, whereas other orders measure it in different contexts.

We sample $R$ orders, each defining a path from $m$ to $x$ that restores the units in that order. Along each path, we numerically integrate gradients and sum the resulting attributions over the feature channels of each snippet, obtaining one value per input step, $a_t^{\pi}$. Averaging these maps over the sampled orders gives $a^{\mathrm{path}}$. Since each path connects $m$ to $x$, the resulting map satisfies completeness, $\sum_t a_t^{\mathrm{path}}=F(x)-F(m)$, up to integration error.

\subsubsection{Approximations for Efficiency}
\label{sec:path-approx}
Evaluating \cref{eq:shapley-order} exactly is expensive: for $N$ units, each order requires $N+1$ forward passes, and $N$ can reach thousands when individual video steps are treated as units~\cite[Remark~5]{sundararajan2017axiomatic}.
With $R$ sampled orders, exact evaluation thus takes $\mathcal{O}(RN)$ forward passes per candidate. To make order averaging practical, we propose two approximations.

\paragraph{Approximation 1: receptive-field window.}
A fixed candidate $q=(l,p,c)$ depends mainly on the input steps within its receptive field at level $l$. We restrict independent ordering to a window centered at the corresponding input-step position that approximates this receptive field:
\begin{equation}
W_q=\bigl\{t:\bigl|t+\tfrac12-(p+\tfrac12)s_l\bigr|\le h s_l\bigr\},
\label{eq:path-window}
\end{equation}
where $t+\tfrac12$ is the center of input step $t$, $(p+\tfrac12)s_l$ is the center of pyramid position $p$ on the input-step axis, and $h$ is the window half-width in level-$l$ positions, chosen to cover this receptive field (\cref{fig:path-units}b; values selection in Suppl.~A.2). For example, in \cref{fig:path-units}b, $q$ is at position $p=2$ of level $l=2$ ($s_l=4$); with $h=1$, \cref{eq:path-window} keeps the steps with $|t+\tfrac12-10|\le4$, that is, $W_q=\{6,\ldots,13\}$.

All steps outside the window form a single \emph{context unit}, restored together at one position in each sampled order rather than excluded from attribution. Each step still receives an attribution, so the map covers the full input, while the number of ordered units, $N=|W_q|+1$, grows with the pyramid level rather than with the video length.

\paragraph{Approximation 2: soft random-order path.}
At coarser pyramid levels, the window can contain many input steps, making exact order evaluation expensive. We address this by allowing consecutive restoration transitions to overlap, approximating each sampled order with a continuous gradient path whose cost depends on the number of integration nodes rather than the number of ordered units (\cref{fig:path-units}c).

For a sampled order $\pi$, let $\kappa_t=r/(N-1)$ denote the normalized rank of input step $t$, where $r\in\{0,\ldots,N-1\}$; steps in the context unit share its rank. At path position $u\in[0,1]$, step $t$ has moved from $m$ toward $x_t$ by the fraction $\beta_t(u)$, where
\begin{equation}
\beta_t(u)=\operatorname{clip}\Bigl(\frac{u-(1-w)\kappa_t}{w},0,1\Bigr),
\label{eq:path-schedule}
\end{equation}
and $w\in(0,1]$ controls the overlap between restoration transitions. 
For example, in \cref{fig:path-units}c ($w=0.3$, $N=9$), rank $r$ starts at $u=(1-w)\,r/8=0.0875\,r$; at $u=0.35$, rank 2 thus has $\beta_t=(0.35-0.175)/0.3\approx0.58$, and ranks 0--3 have $\beta_t=1$, $0.875$, $0.58$, and $0.29$ while the others have not started, so consecutive ranks overlap.
This overlap schedule is the same as in Spectral IG~\cite{kim2026spectral}, but there it follows a single order of singular components, whereas here it follows the rank in a sampled restoration order. At $w=1$, all steps change together, recovering straight-line IG; as $w\rightarrow0$, the path approaches one-at-a-time restoration, so averaging over sampled orders approaches the order average of \cref{eq:shapley-order}. 

\paragraph{Hybrid:} For each candidate, we use the faster of the soft path and exact forward differences (Suppl.~A.4). Exact evaluation requires $|W_q|+2$ forward passes per order, whereas the soft path uses $M+1$ gradient evaluations for a fixed number $M$ of integration intervals, making the former preferable for narrow windows and the latter for wider ones (\cref{fig:path-units}d).

\subsection{Boundary-Specific Path and Baseline}
\label{sec:boundary-map}

Order averaging applies to all three targets, but boundary localization requires a different attribution design from classification. 
Unlike a class score that aggregates discriminative evidence, boundary localization depends on where evidence changes along the temporal sequence.
We therefore tailor both the integration path and the reference to boundary placement (\cref{fig:base-trunc}).

\subsubsection{Boundary Path}
\label{sec:boundary-path}

Boundary attribution is particularly sensitive to how temporal evidence is organized along restoration paths. In partially restored inputs, transitions between present and absent snippets may influence boundary predictions, consistent with the greater variation of boundary attributions across sampled orders than class attributions (Suppl.~C.1). 
We therefore structure restoration orders around the detector's temporal cells: groups of about $s_l$ consecutive input steps around each position at the candidate's pyramid level, where both detectors predict boundary offsets~\cite{shi2023tridet,zhang2022actionformer}.

Formally, the cells within the window are
\begin{equation}
C_j=\bigl\{t\in W_q:\lfloor (t+s_l/2)/s_l\rfloor=j\bigr\},
\label{eq:path-cell}
\end{equation}
i.e., each step is assigned to its nearest anchor $j s_l$ at level $l$, the point from which both detectors predict boundary offsets.
We then sample restoration orders hierarchically: first ordering the cells and the context unit randomly, then ordering the steps within each cell. This keeps steps from the same cell consecutive while preserving step-level attributions, so artificial transitions are largely confined to cell boundaries or the cell being restored (\cref{fig:base-trunc}a).
For example, in \cref{fig:base-trunc}a, \cref{eq:path-cell} splits $W_q=\{6,\ldots,13\}$ into $C_j=\{6,\ldots,9\}$ and $C_{j+1}=\{10,\ldots,13\}$. After the units with $r\le3$ are restored, $\pi_1$ has restored steps 9, 11, and 12, so restored and unrestored steps meet at four places inside $W_q$; the cell-structured order has restored all of $C_{j+1}$, so they meet only at the cell border.

In the non-overlapping limit, this construction corresponds to the Owen value~\cite{owen1977values} with the cells and the context unit as a priori unions.

\subsubsection{Boundary Baseline}
\label{sec:boundary-baseline}

The choice of reference determines what a prediction is explained against~\cite{sundararajan2020many}, particularly for regression outputs~\cite{letzgus2022toward}. For boundary attribution, we seek to identify the evidence that places the start or end of an existing candidate given the surrounding video. However, the global mean baseline $m$ removes all input evidence, so attribution relative to it also reflects the transition from the candidate's default boundary predictions to those of the original video. We therefore introduce boundary-anchored references that preserve the surrounding evidence while selectively removing content adjacent to the predicted boundary.

For each boundary $\tau\in\{\mathrm{s},\mathrm{e}\}$, we define nested masked spans just inside the predicted interval:
\[
H_k^{\mathrm{s}}=[t_q^{\mathrm{s}},t_q^{\mathrm{s}}+\ell_k),
\qquad
H_k^{\mathrm{e}}=[t_q^{\mathrm{e}}-\ell_k,t_q^{\mathrm{e}}),
\]
where $\ell_k=\max(1,\operatorname{round}(\rho_k|I_q|))$ with $(\rho_1,\rho_2,\rho_3)=(1/6,1/3,1/2)$. 
The reference $b^{(\tau,k)}$ equals $x$ outside $H_k^\tau$ and takes the dataset mean within it, adapting the masked width to the predicted action duration without using ground-truth information.
For example, in \cref{fig:base-trunc}b, $|I_q|=6$ gives $\ell_k=1,2,3$, so $b^{(\mathrm{e},1)}$ replaces only step 12, whereas $b^{(\mathrm{e},3)}$ replaces steps 10 to 12.
Straight-line IG from each reference $b^{(\tau,k)}$ to $x$ produces a local attribution map $a^{(\tau,k)}$, supported only on the masked span $H_k^\tau$.

\paragraph{Fusion.}
For boundary targets, the cell-structured path map distributes the boundary
change relative to $m$ over the whole input, whereas the boundary-anchored
maps probe the evidence next to the predicted boundary.
We average the two wherever both are available (\cref{fig:base-trunc}c):
\begin{equation}
a_t^\tau
=
\frac{
\lambda a_t^{\mathrm{path},\tau}
+
\sum_k
\mathbf{1}[t\in H_k^\tau]
a_t^{(\tau,k)}
}{
\lambda+
\sum_k\mathbf{1}[t\in H_k^\tau]
},
\qquad
\lambda=1 .
\label{eq:combine-map}
\end{equation}
For example, in \cref{fig:base-trunc}c, step 12 lies in all three end spans and receives $(a^{\mathrm{path},\mathrm{e}}_{12}+\sum_k a^{(\mathrm{e},k)}_{12})/4$, whereas step 10 lies only in $H_3^{\mathrm{e}}$ and receives $(a^{\mathrm{path},\mathrm{e}}_{10}+a^{(\mathrm{e},3)}_{10})/2$.
This step-wise normalization preserves the global path attribution outside the masked spans. The fused map is not complete with respect to a single reference, though each individual map is.

\begin{table*}[t]
  \centering
  \footnotesize
  \setlength{\tabcolsep}{1.65pt}
  \begin{tabular}{@{}l<{\hspace{-1.3pt}}cc<{\hspace{0.7pt}} cccc cccc cc<{\hspace{1.7pt}} cccc cccc cc@{}}
    \toprule
    & & & \multicolumn{10}{c}{TriDet} & \multicolumn{10}{c}{ActionFormer} \\
    \cmidrule(r{5pt}){4-13}\cmidrule(l){14-23}
    & & & \multicolumn{4}{c}{THUMOS14} & \multicolumn{4}{c}{EPIC-KITCHENS} & \multicolumn{2}{c}{Hours $\downarrow$} & \multicolumn{4}{c}{THUMOS14} & \multicolumn{4}{c}{EPIC-KITCHENS} & \multicolumn{2}{c}{Hours $\downarrow$} \\
    \cmidrule(lr){4-7}\cmidrule(lr){8-11}\cmidrule(r{5pt}){12-13}\cmidrule(lr){14-17}\cmidrule(lr){18-21}\cmidrule(l){22-23}
    & & & & & \multicolumn{2}{c}{Sens.-$n$\,$\uparrow$} & & & \multicolumn{2}{c}{Sens.-$n$\,$\uparrow$} & & & & & \multicolumn{2}{c}{Sens.-$n$\,$\uparrow$} & & & \multicolumn{2}{c}{Sens.-$n$\,$\uparrow$} & & \\
    \cmidrule(lr){6-7}\cmidrule(lr){10-11}\cmidrule(lr){16-17}\cmidrule(lr){20-21}
    Method & Win. & Path & Del.$\uparrow$ & Ins.$\uparrow$ & rem. & rest. & Del.$\uparrow$ & Ins.$\uparrow$ & rem. & rest. & TH & EP & Del.$\uparrow$ & Ins.$\uparrow$ & rem. & rest. & Del.$\uparrow$ & Ins.$\uparrow$ & rem. & rest. & TH & EP \\
    \midrule
    \multicolumn{23}{@{}l}{\emph{Single path}} \\
    IG~\cite{sundararajan2017axiomatic} &  &  & 61.2 & 39.8 & 62.4 & 78.0 & 50.3 & 34.3 & 62.1 & 64.4 & -- & -- & 63.1 & 38.9 & 66.7 & 81.8 & 47.6 & 35.2 & 59.3 & 62.9 & -- & -- \\
    Guided IG~\cite{kapishnikov2021guided} &  &  & 61.3 & 39.8 & 62.4 & 78.1 & 50.4 & 34.5 & 62.2 & 64.5 & -- & -- & 63.2 & 38.9 & 66.7 & 81.8 & 47.6 & 35.2 & 59.5 & 63.1 & -- & -- \\
    Adaptive nodes~\cite{nakai2026adaptive} &  &  & 61.2 & 39.8 & 62.4 & 78.1 & 50.3 & 34.4 & 62.2 & 64.5 & -- & -- & 63.2 & 38.9 & 66.7 & 81.8 & 47.6 & 35.2 & 59.4 & 63.0 & -- & -- \\
    Left to right$^{\ddagger}$ &  &  & 58.4 & 36.6 & 55.5 & 69.8 & 50.6 & 35.0 & 63.1 & 66.2 & -- & -- & 59.5 & 35.1 & 60.0 & 74.2 & 48.4 & 35.1 & 61.7 & 64.9 & -- & -- \\
    Right to left$^{\ddagger}$ &  &  & 58.0 & 37.3 & 58.4 & 70.5 & 49.2 & 34.1 & 62.4 & 65.9 & -- & -- & 59.2 & 35.9 & 60.1 & 74.1 & 47.0 & 35.1 & 61.0 & 65.1 & -- & -- \\
    \midrule
    \multicolumn{23}{@{}l}{\emph{Other attributions}} \\
    TIMING~\cite{jang2025timing}$^{\dagger}$ &  &  & 57.4 & 37.5 & 62.1 & 69.9 & 35.1 & 23.0 & 41.0 & 40.3 & -- & -- & 59.2 & 36.4 & 63.8 & 72.5 & 33.7 & 23.6 & 40.5 & 40.8 & -- & -- \\
    \midrule
    \multicolumn{23}{@{}l}{\emph{Average over random orders}} \\
       &  & soft & 63.0 & 43.2 & 68.3 & \textbf{81.4} & 56.3 & 41.3 & 71.1 & 73.7 & 3.87 & 5.25 & 64.5 & 41.0 & 70.6 & \textbf{84.8} & 52.9 & 40.8 & 68.7 & 71.7 & 6.16 & 7.62 \\
       &  & exact & 62.7 & 43.4 & \textbf{68.4} & 81.3 & \textbf{56.4} & 41.3 & \textbf{71.4} & \textbf{74.0} & 36.85 & 12.50 & 64.4 & 41.0 & \textbf{70.7} & 84.7 & \textbf{53.0} & 40.8 & 69.0 & \textbf{72.0} & 49.20 & 17.09 \\
       & \checkmark & soft & \textbf{63.3} & \textbf{43.8} & 68.2 & 81.2 & 56.1 & 41.3 & 70.9 & 73.5 & 3.87 & 5.25 & \textbf{64.6} & 41.1 & 70.6 & 84.6 & 52.9 & 40.9 & 68.8 & 71.6 & 6.16 & 7.62 \\
       & \checkmark & exact & 62.9 & 43.7 & 68.3 & 81.1 & 56.2 & \textbf{41.5} & 71.1 & 73.7 & 5.99 & 3.42 & 64.5 & \textbf{41.2} & \textbf{70.7} & 84.5 & 52.9 & \textbf{41.1} & \textbf{69.1} & 71.8 & 10.19 & 4.10 \\
    \rowcolor{gray!12}
    Ours & \checkmark & hybrid & 63.1 & \textbf{43.8} & 68.3 & 81.2 & 56.2 & 41.4 & 71.0 & 73.6 & 3.03 & 2.60 & 64.5 & \textbf{41.2} & 70.6 & 84.6 & 52.9 & 41.0 & 69.0 & 71.8 & 4.77 & 3.41 \\
    \bottomrule
  \end{tabular}
    \caption{\textbf{Class-score faithfulness and cost} (metrics $\times100$). Random-order averaging is more faithful than IG on every metric for both detectors and datasets; the hybrid keeps this faithfulness at the lowest cost. Del., Ins.: deletion drop and insertion gain over the first $L$ input steps; Sens.-$n$: block Sensitivity-$n$ under removal (rem.) and restoration (rest.). Win.: ordering restricted to the receptive-field window; Path: soft path, exact forward differences, or hybrid, the faster of the two per candidate. Hours: class maps summed over all candidates on THUMOS14 (TH) and EPIC-KITCHENS-100 (EP), measured as in the supplementary material. $^{\dagger}$TIMING: 269 of the 272 TriDet and 273 ActionFormer EPIC candidates (two videos are shorter than its segments). $^{\ddagger}$One order of the window units, restored one at a time. --: not timed. Shaded: ours. Bold: best in each column except hours.}

  \label{tab:exp-class}
\end{table*}

\begin{table*}[t]
  \centering
  \footnotesize
  \setlength{\tabcolsep}{1.7pt}
  \begin{tabular}{@{}l<{\hspace{-1.4pt}}cccc<{\hspace{-0.4pt}} ccc ccc cc<{\hspace{1.6pt}} ccc ccc cc@{}}
    \toprule
    & & & & & \multicolumn{8}{c}{TriDet} & \multicolumn{8}{c}{ActionFormer} \\
    \cmidrule(r{5pt}){6-13}\cmidrule(l){14-21}
    & & & & & \multicolumn{3}{c}{THUMOS14} & \multicolumn{3}{c}{EPIC-KITCHENS} & \multicolumn{2}{c}{Hours $\downarrow$} & \multicolumn{3}{c}{THUMOS14} & \multicolumn{3}{c}{EPIC-KITCHENS} & \multicolumn{2}{c}{Hours $\downarrow$} \\
    \cmidrule(lr){6-8}\cmidrule(lr){9-11}\cmidrule(r{5pt}){12-13}\cmidrule(lr){14-16}\cmidrule(lr){17-19}\cmidrule(l){20-21}
    & & & & & Ins.-- & \multicolumn{2}{c}{Sens.-$n$\,$\uparrow$} & Ins.-- & \multicolumn{2}{c}{Sens.-$n$\,$\uparrow$} & & & Ins.-- & \multicolumn{2}{c}{Sens.-$n$\,$\uparrow$} & Ins.-- & \multicolumn{2}{c}{Sens.-$n$\,$\uparrow$} & & \\
    \cmidrule(lr){7-8}\cmidrule(lr){10-11}\cmidrule(lr){15-16}\cmidrule(lr){18-19}
    Method & Win. & Cells & Refs. & Path & Del.$\uparrow$ & rem. & rest. & Del.$\uparrow$ & rem. & rest. & TH & EP & Del.$\uparrow$ & rem. & rest. & Del.$\uparrow$ & rem. & rest. & TH & EP \\
    \midrule
    \multicolumn{21}{@{}l}{\emph{Single path}} \\
    IG~\cite{sundararajan2017axiomatic} &  &  &  &  & 22.2 & 35.6 & 43.2 & 15.0 & 13.6 & 25.6 & -- & -- & 22.2 & 43.4 & 44.1 & 5.5 & $-$2.4 & 7.6 & -- & -- \\
    Guided IG~\cite{kapishnikov2021guided} &  &  &  &  & \underline{22.4} & 36.1 & \underline{43.7} & \underline{15.2} & 14.0 & \underline{25.9} & -- & -- & \underline{22.4} & 43.9 & \underline{44.7} & 5.5 & $-$2.3 & 7.7 & -- & -- \\
    Adaptive nodes~\cite{nakai2026adaptive} &  &  &  &  & 22.2 & 35.6 & 43.3 & 14.9 & 13.3 & 25.4 & -- & -- & 22.2 & 43.4 & 44.3 & 5.4 & $-$2.5 & 7.7 & -- & -- \\
    Left to right$^{\ddagger}$ &  &  &  &  & 15.3 & 22.0 & 26.3 & 12.9 & 17.0 & 14.9 & -- & -- & 16.6 & 35.8 & 33.1 & \underline{11.6} & 15.3 & 7.4 & -- & -- \\
    Right to left$^{\ddagger}$ &  &  &  &  & 15.7 & 22.9 & 28.5 & 11.5 & 12.9 & 12.6 & -- & -- & 15.4 & 30.1 & 31.1 & 11.3 & 13.4 & 8.7 & -- & -- \\
    \midrule
    \multicolumn{21}{@{}l}{\emph{Other attributions}} \\
    TIMING~\cite{jang2025timing}$^{\dagger}$ &  &  &  &  & 22.1 & \underline{43.9} & 39.0 & 11.7 & \underline{22.1} & 19.4 & -- & -- & 20.9 & \underline{45.1} & 38.8 & 7.9 & \underline{17.9} & \underline{10.6} & -- & -- \\
    \midrule
    \multicolumn{21}{@{}l}{\emph{Average over random orders}} \\
          &  &  &  & soft & 25.1 & 38.2 & 48.2 & 19.6 & 15.7 & 31.1 & 5.05 & 8.87 & \textbf{25.3} & 46.4 & 47.6 & 9.2 & $-$0.5 & 11.1 & 9.80 & 12.25 \\
          &  &  &  & exact & 25.0 & 38.3 & 48.4 & 20.8 & 18.3 & 32.4 & 36.85 & 12.50 & 25.2 & 46.5 & 47.5 & 10.9 & 2.7 & 12.4 & 49.20 & 17.09 \\
          & \checkmark &  &  & soft & 25.0 & 38.2 & 48.1 & 19.6 & 15.1 & 30.9 & 5.05 & 8.87 & 25.2 & 46.5 & 47.6 & 9.1 & $-$0.9 & 10.8 & 9.80 & 12.25 \\
          & \checkmark &  &  & exact & 25.1 & 38.4 & 48.5 & 20.9 & 17.6 & 32.1 & 5.99 & 3.42 & 25.1 & 46.6 & 47.5 & 10.8 & 2.1 & 12.1 & 10.19 & 4.10 \\
          & \checkmark &  &  & hybrid & 25.1 & 38.2 & 48.2 & 20.3 & 16.8 & 31.5 & 3.62 & 3.05 & 25.2 & 46.5 & 47.6 & 10.6 & 1.8 & 12.0 & 6.37 & 3.95 \\
          & \checkmark & \checkmark &  & hybrid & 25.5 & 40.9 & \textbf{49.7} & 21.2 & 21.6 & 33.0 & 3.62 & 3.05 & 25.2 & 47.9 & \textbf{48.6} & 12.6 & 5.8 & 12.4 & 6.37 & 3.95 \\
          & \checkmark &  & \checkmark & hybrid & 25.8 & 42.9 & 47.0 & 23.3 & 29.3 & 35.1 & 5.69 & 4.28 & 25.1 & 48.5 & 45.3 & 15.0 & 20.3 & 16.3 & 7.91 & 5.75 \\
    \rowcolor{gray!12}
    Ours & \checkmark & \checkmark & \checkmark & hybrid & \textbf{26.0} & \textbf{44.4} & 47.6 & \textbf{24.3} & \textbf{34.3} & \textbf{36.7} & 5.69 & 4.28 & 25.0 & \textbf{49.5} & 46.2 & \textbf{16.2} & \textbf{25.7} & \textbf{17.7} & 7.91 & 5.75 \\
    \bottomrule
  \end{tabular}
    \caption{\textbf{Boundary faithfulness and cost} (metrics $\times100$, averaged over start and end; other abbreviations and symbols as in \cref{tab:exp-class}). The full boundary map is more faithful than every compared method on every metric. Ins.--Del.: insertion--deletion area; Sens.-$n$ uses the signed boundary displacement. Cells: cell-structured orders; Refs.: fused with the boundary-anchored references. Hours: start and end maps, including the reference attributions where used; exact runs also give the class map, so their hours match \cref{tab:exp-class}. Underline: best compared method.}
    
  \label{tab:exp-boundary}
\end{table*}

\section{Experiments}
\label{sec:experiments}

\subsection{Experimental Setup}
\label{sec:exp-setup}

\paragraph{Data and candidates.}
We evaluate TriDet~\cite{shi2023tridet} and ActionFormer~\cite{zhang2022actionformer} on THUMOS14~\cite{idrees2017thumos} and the verb task of EPIC-KITCHENS-100~\cite{damen2022rescaling}.
We select correctly localized final detections from the THUMOS14 test and EPIC-KITCHENS-100 validation splits, requiring a temporal IoU of at least 0.7 with the ground truth. Each detection is attributed through its pre-NMS candidate. Sampling across pyramid levels (Suppl.~B.1) gives 199 and 272 TriDet and 198 and 273 ActionFormer candidates on the two datasets, respectively.

\paragraph{Faithfulness metrics.}
We measure faithfulness with the following perturbation-based metrics.

\emph{Deletion and insertion}~\cite{petsiuk2018rise} evaluate class attribution by removing or restoring steps in descending attribution order. 
A faithful map should produce a sharp class-score drop during deletion and a rapid rise during insertion.

\emph{Insertion--deletion area} evaluates boundary attribution with the regression versions of deletion and insertion by Hama \etal~\cite{hama2023deletion}, as a boundary is a regressed coordinate that evidence can move earlier or later. Following them, we restore steps in descending and ascending attribution order and use the signed area between the two curves, which has zero expectation under random rankings even when inputs interact. We compute it within $R_q$, the predicted interval extended by its duration on both sides, and normalize it by the predicted duration.

\emph{Block Sensitivity-$n$} extends Sensitivity-$n$~\cite{ancona2018towards} (see also~\cite{bhatt2020evaluating}) from random feature subsets to contiguous temporal blocks, because neighboring snippets can carry evidence that matters jointly. Following segment- and subsequence-level evaluations~\cite{gevaert2022evaluating,schlegel2019towards,simic2025comprehensive}, we use all contiguous blocks $B\subseteq R_q$ of widths $n\in\{1,2,4,8,16,32,|I_q|\}$, with $n\leq|R_q|$.
For each width, we compute the Pearson correlation across blocks between $A(B)=\sum_{t\in B}a_t$ and the output change from removing or restoring $B$, and average over widths with at least three blocks. We report removal and restoration separately. (All metric details are provided in Suppl.~B.1.)

\paragraph{Evaluation and compared methods.}
Perturbation tests replace removed steps with time-aligned features from donor videos whose annotations exclude the explained class, with results averaged over three donors. We compare against straight-line IG~\cite{sundararajan2017axiomatic}, Guided IG~\cite{kapishnikov2021guided}, adaptive node placement from Adaptive IG-ODAM~\cite{nakai2026adaptive}, TIMING~\cite{jang2025timing}, and fixed left-to-right and right-to-left restoration orders. Our maps use $R=32$ orders and $w=0.1$ (selection in Suppl.~C.1; further settings in Suppl.~B.2).

\subsection{Order Averaging versus Single Paths}
\label{sec:exp-faithfulness}

\paragraph{Existing variants of IG bring no consistent gain.}
Guided IG and adaptive node placement keep the path on or near IG's straight line and stay close to IG on every metric for both class and boundary targets (\cref{tab:exp-class,tab:exp-boundary}). 
Fixed restoration orders fall below IG on every metric on THUMOS14 for both targets, whereas on EPIC-KITCHENS-100 left-to-right restoration exceeds IG on most class metrics and in boundary removal Sensitivity-$n$. This may be because contiguous context is more coherent in the unedited recordings of EPIC-KITCHENS-100~\cite{damen2022rescaling} than in the THUMOS14 YouTube videos, which can contain scene changes~\cite{ahn2024interpretable}.
TIMING falls below IG on every class metric. Its default settings leave many steps of short inputs unattributed, most often on EPIC-KITCHENS-100 (Suppl.~B.2). For boundaries, TIMING performs better relative to IG. It attributes each step with most of the video kept at its input values, and this context matters much more for boundaries than for class scores (Suppl.~C.1). The removal test also starts from the clean video, whereas the restoration test starts from a replaced input (\cref{sec:exp-boundary}). Accordingly, TIMING is the strongest compared method in boundary removal Sensitivity-$n$ in every setting but mostly falls below IG in restoration.

\paragraph{Random-order averaging improves on IG in every setting.}
As noted in \cref{sec:intro}, a step's contribution depends on its temporal context: it is larger when its neighbors are absent than when they are present, and a single random order is even less faithful than IG (Suppl.~C.1). Averaging over random orders covers many contexts and improves on IG for every metric, dataset, detector, and target. For class scores, it also surpasses every compared method. For boundaries, random-order averaging is not yet better than TIMING in removal Sensitivity-$n$ in most settings. The boundary-specific path and baseline (\cref{sec:exp-boundary}) close this gap and surpass TIMING in every setting.

\subsection{Approximations and Cost}
\label{sec:exp-class}

\paragraph{Both approximations preserve faithfulness.}
For both targets, the receptive-field window changes every metric by less than one point (\cref{tab:exp-class,tab:exp-boundary}). ActionFormer's input gradients vanish outside the window, and for TriDet a single context unit suffices for the influence from outside it (Suppl.~A.2). Soft paths also stay close to exact differences, as they keep the restoration order and thus the averaging over contexts. The exception is boundaries on EPIC-KITCHENS-100, where exact differences are 2.5--3.0 points higher in removal Sensitivity-$n$. This gap is concentrated at pyramid levels 3--5, where wide windows make many units change at once along the soft path (Suppl.~C.3).

\paragraph{The hybrid takes the faster method per candidate.}
The window makes exact differences 3.7--6.1 times faster. Soft paths have a fixed cost per order, making them faster for wide windows and exact differences faster for narrow ones. The hybrid selects the faster method per candidate, achieving a 4.1--12.2$\times$ speedup over exact differences without the window while staying within 1.5 points on every metric.

\subsection{Boundary-Specific Path and Baseline}
\label{sec:exp-boundary}
\label{sec:exp-references}

\paragraph{Cell-structured orders improve boundary maps, most for wide cells.}
Cell-structured orders raise removal and restoration Sensitivity-$n$ for both detectors and datasets, with or without the references, most on EPIC-KITCHENS-100 (\cref{tab:exp-boundary}). Decomposing this gain by pyramid level (Suppl.~C.3) shows that it is small at levels 0--1, where cells hold one or two steps, and largest at levels 3--5. This is consistent with the motivation of cells (\cref{sec:boundary-path}): they confine transitions between restored and replaced steps to cell borders, which matters more as cells grow wider.

\paragraph{Boundary-anchored references improve removal on both datasets.}
The references raise removal Sensitivity-$n$ on both datasets. On EPIC-KITCHENS-100, they improve every metric. On THUMOS14, they lower restoration Sensitivity-$n$.  
This is consistent with the input states used for attribution: reference maps perturb mostly clean inputs, like removal, whereas path maps start from mostly replaced inputs, like restoration.
We compare boundary spans with random spans of equal widths (Suppl.~C.5). 
Boundary spans perform better on every metric, so the placement of the spans matters, not only the masking itself.

\paragraph{The full boundary map outperforms all compared methods.}
It surpasses every compared method on every metric and gives the highest removal Sensitivity-$n$ in every setting. On EPIC-KITCHENS-100, it also improves on each of its components on every metric. On THUMOS14, restoration Sensitivity-$n$ stays below that of cell-structured orders alone, as the references lower it. 
A larger fusion weight $\lambda$ on the path map reduces this loss but also lowers removal Sensitivity-$n$ (Suppl.~C.5). 
The insertion--deletion area changes by less than one point across variants. Both metrics restore steps into a replaced input, consistent with the input-state argument above.

\section{Analysis}
\label{sec:analysis}
\label{sec:analysis-rule}

\Cref{sec:intro} noted that predictions with localization errors can still be among the highest-scored, which detector designs attribute to a classifier relying on content that is less informative about boundary placement.
We test this explanation at the level of input evidence by asking whether the class score of a detection tracks its boundaries' response to the content that places them.

\subsection{Setup}
\label{sec:analysis-setup}
We analyze TriDet on THUMOS14 (TH) and EPIC-KITCHENS-100 (EP), TriDet retrained with InternVideo2-6B features~\cite{wang2024internvideo2} on THUMOS14 (IV, 3.9 mAP points higher), and ActionFormer on THUMOS14 (AF).
The cases are the candidates of \cref{sec:exp-setup} and the maps are the windowed class map and the full boundary map (\cref{sec:path,sec:boundary-map}).
All analyses concern the $k\approx|I|/6$ steps just inside or just outside a predicted boundary.

\begin{figure}[t]
  \centering
  \includegraphics[width=\columnwidth]{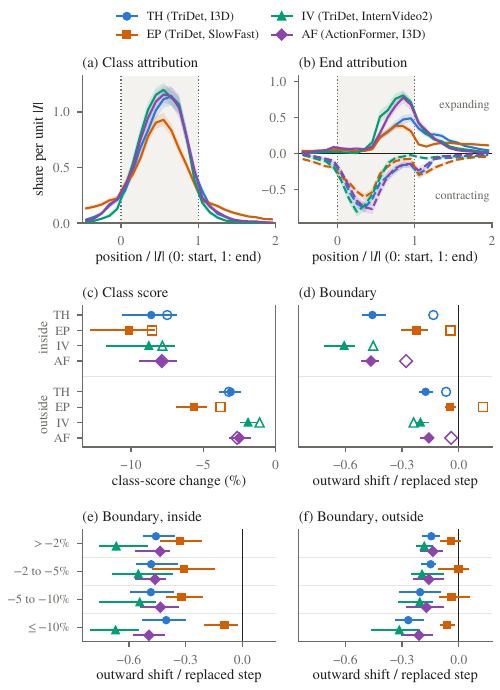}
  \caption{\textbf{Class and boundary evidence at the boundary.}
  The class-score change under removal indicates little about boundary shift.
  (a) Mean positive class attribution and (b) mean signed end-boundary attribution (solid: expanding, outward; dashed: contracting, inward), as a share of the map's total absolute attribution per unit of predicted duration.
  (c, d) Replacing the $k$ steps just inside or outside a boundary with background: relative change of the class score (c) and outward shift of the boundary per replaced step (d); filled markers: measured; open markers: the change the maps attribute to the replaced steps (class map; boundary map without references).
  (e, f) The shift of (d), inside (e) and outside (f), for boundaries grouped by the relative change of their class score.}
  \label{fig:analysis}
\end{figure}

\subsection{Class and Boundary Evidence}
\label{sec:analysis-results}

\paragraph{The maps place class and boundary evidence differently.}
A boundary's expanding (outward) attribution concentrates in the half of the box next to it and just beyond (\cref{fig:analysis}b), with its strongest step mostly at or just inside the boundary's own pyramid cell, whereas its contracting (inward) attribution is stronger in the far half of the box than in the near half.
These patterns hold without the boundary-anchored references and for the start and end separately.
In contrast, class attribution (\cref{fig:analysis}a) peaks inside the box and declines toward the boundary, yet it is still positive at the boundary's cell in most cases and often one cell beyond.
The $k$ steps just inside a boundary thus carry much of the expanding support in the half next to it but only about half of an even share of the class score, and the steps just beyond it carry little of the score.
Even across individual boundaries, the class and boundary attributions of these steps are at most weakly related (Spearman $-0.17$ to $0.18$, inside and outside; supplementary material).
The maps therefore suggest that the content placing a boundary carries comparatively little of the class score.

\paragraph{Removal confirms the separation.}
We remove these steps by replacing them with real background features from another video of the same split (supplementary material).
In the median, removal just inside a boundary lowers the class score and retracts the boundary, whereas removal just outside changes both outputs less (\cref{fig:analysis}c, d). Across boundaries, we compare the measured changes with those the maps attribute to the replaced steps (minus their summed attribution) by Spearman correlation.

The class map ranks the change in the class score (0.46--0.75) but not the boundary shift ($-0.27$ to $0.07$), and neither does the measured score change ($-0.30$ to $0.08$): large score drops come with no larger retractions inside the box and with at most slightly larger ones outside it (\cref{fig:analysis}e, f).
The boundary map, in contrast, ranks the shift (0.21--0.53).

\paragraph{Implications.}
First, in both detectors, the class score responds to boundary-adjacent content, but its change indicates little about how far the boundary moves when this content is removed.
This is consistent with the observation that predictions with localization errors can still be among the highest-scored~\cite[Fig.~5]{alwassel2018diagnosing}, and our analysis provides input-level evidence from removal interventions on fixed candidates supporting the explanation that classification relies on content that is less informative about boundary placement~\cite{zhao2017ssn,liu2022tadtr,hu2022estimation}.
Second, since the class map does not capture the boundary's response, faithfully explaining localization requires attributing the boundary coordinates themselves rather than relying on the class map.

\section{Conclusion}

We introduced a structure-aware attribution framework for TAL that makes order averaging computationally practical and tailors attribution to temporal boundaries. 
Across two datasets and two detectors, our faithfulness evaluation, direction-aware for boundaries, shows improved attribution faithfulness over existing methods for both classification and boundary predictions.
Attribution analysis and controlled interventions further show that class-score changes do not reliably track boundary displacement, highlighting the importance of explaining localization separately from classification.

\paragraph{Limitations and future work.}
Our analysis focuses on correctly localized detections from two dense TAL detectors and does not directly establish the causes of high-confidence localization errors. Future work could extend the analysis to localization failures and investigate the internal mechanisms behind classification and boundary responses, for example through activation patching.
{
    \small
    \bibliographystyle{ieeenat_fullname}
    \bibliography{main}
}

\clearpage
\setcounter{table}{0}
\setcounter{figure}{0}
\setcounter{equation}{0}
\renewcommand{\thetable}{S\arabic{table}}
\renewcommand{\thefigure}{S\arabic{figure}}
\renewcommand{\theequation}{S\arabic{equation}}
\renewcommand{\theHtable}{S\arabic{table}}
\renewcommand{\theHfigure}{S\arabic{figure}}
\renewcommand{\theHequation}{S\arabic{equation}}
\setcounter{dbltopnumber}{3}
\maketitlesupplementary
\appendix
\section{Method Details}
\label{sec:app-method}

\subsection{Problem Formulation}
\label{sec:app-prelim}
\paragraph{Predicted interval.}
The clean predicted interval $I_q$ contains the input steps whose time on the detector's feature grid lies in $[F_q^{\mathrm{s}},F_q^{\mathrm{e}})$, cut to the input.

\subsection{Random-Order Path Attribution for TAL}
\label{sec:app-path}

\subsubsection{Averaging over Temporal Contexts}
\label{sec:app-path-avg}

The $R$ orders are sampled in antithetic pairs, pairing each random order with its reverse~\cite{mitchell2022sampling}, which reduces the variance of the order average.

\subsubsection{Approximations for Efficiency}
\label{sec:app-path-approx}

\paragraph{Receptive-field window.}
We set the half-width $h$ of the receptive-field window from each detector's architecture.
For ActionFormer, we use $h=26$ on THUMOS14 and $h=16$ on EPIC-KITCHENS-100, which covers its full receptive field at every pyramid level: its backbone uses local attention and its heads are convolutional, so input gradients are exactly zero outside the window.
We verified this on 18 candidates per dataset: the farthest input step with a nonzero gradient lies at most 25 and 15 cells from the candidate, respectively, and replacing all steps outside the window with $m$ leaves every output unchanged up to floating-point error.
For TriDet, we use $h=20$, which covers its local receptive field: the 16 positions on either side read by its localization head and the reach of its convolutional heads.
Its backbone additionally contains a global average-pooling branch~\cite{shi2023tridet}, through which steps outside the window also affect the candidate; this influence enters through the context unit.
With exact differences, the contribution of the context unit is split equally among its steps; along the soft path, each of its steps receives its own gradient attribution.

\paragraph{Soft-path integration.}
We integrate each soft path using the trapezoidal rule at $u_i=i/M$, $i=0,\ldots,M$.
With $z_i$ the input at $u_i$ on the soft path of order $\pi$ (Eq.~(3) of the main paper) and $e_i=\frac12(z_{i+1}-z_{i-1})$, where $z_{-1}=z_0$ and $z_{M+1}=z_M$, the attribution along one path is
\begin{equation}
A^\pi
\approx
\sum_{i=0}^{M}
\nabla F(z_i)\odot e_i.
\end{equation}
The final map averages these attributions over the
$R$ sampled orders.
We use $M=100$ integration intervals, giving
approximately ten intervals per restoration ramp
at $w=0.1$.

\subsection{Boundary-Specific Path and Baseline}
\label{sec:app-boundary}

\subsubsection{Boundary Path}
\label{sec:app-boundary-path}
Eq.~(4) of the main paper centers cell $j$ on input step $j s_l$, the anchor point at which both detectors place position $j$ and from which they regress boundary offsets; the window of Eq.~(2) is centered on the middle of the span of position $p$, $(p+\frac12)s_l$.
The start and end maps share their sampled orders and, with exact differences, their forward passes.
These orders are also sampled in antithetic pairs (\cref{sec:app-path-avg}); reversing a cell-structured order keeps the steps of each cell consecutive, so both orders of a pair are cell-structured.

\subsubsection{Boundary Baseline}
\label{sec:app-boundary-baseline}
Each local map $a^{(\tau,k)}$ integrates straight-line IG from $b^{(\tau,k)}$ to $x$ with 50 right Riemann steps.

\subsection{Complete Procedure}
\label{sec:app-algorithm}
\Cref{alg:app-maps} summarizes how the class map and the two boundary maps of one candidate are computed; equation numbers refer to the main paper.
The two calls of \textsc{PathMap} use different restoration orders: uniform random orders of the units for the class score, and cell-structured orders for the start and end (\cref{sec:app-boundary-path}).
Every per-order map is summed over feature channels; with exact differences, the context unit's share is split equally among its steps (\cref{sec:app-path-approx}). The costs $c_{\mathrm{fwd}}$ of a forward pass and $c_{\mathrm{grad}}(\mathcal{F})$ of a gradient evaluation of the targets $\mathcal{F}$ are fitted to measured times (\cref{sec:app-runtime}), so the class and boundary maps of a candidate can use different methods.

\begin{algorithm}[t]
\caption{Attribution maps of a candidate $q$}
\label{alg:app-maps}
\small
\begin{algorithmic}[1]
\Require input $x\in\mathbb{R}^{T\times D}$, mean baseline $m$, candidate $q=(l,p,c)$ with outputs $F_q^{\mathrm{cls}},F_q^{\mathrm{s}},F_q^{\mathrm{e}}$; $R$, $w$, $M$, $h$, $(\rho_k)_{k=1}^{3}$, $\lambda$
\Ensure maps $a^{\mathrm{cls}},a^{\mathrm{s}},a^{\mathrm{e}}\in\mathbb{R}^{T}$
\State $W_q\gets$ window of $q$ (Eq.~(2)); other steps: context unit
\State $C_j\gets$ cells of $W_q$ (Eq.~(4))
\Function{PathMap}{order sampler, targets $\mathcal{F}$}
  \State $\mathit{exact}\gets R(|W_q|+2)\,c_{\mathrm{fwd}}<(R(M+1)+2)\,c_{\mathrm{grad}}(\mathcal{F})$
  \For{$i=1,\ldots,R/2$}
    \State draw an order $\pi$ of the $|W_q|+1$ units
    \For{$\pi'\in\{\pi,\operatorname{reverse}(\pi)\}$} \Comment{antithetic pair}
      \If{$\mathit{exact}$}
        \State $a^{\pi'}_F\gets$ forward differences along $\pi'$
      \Else
        \State $a^{\pi'}_F\gets$ soft-path IG (Eq.~(3)), $M$ intervals
      \EndIf
    \EndFor
  \EndFor
  \State \Return $a^{\mathrm{path}}_F\gets$ mean of the $a^{\pi'}_F$, $F\in\mathcal{F}$
\EndFunction
\State $a^{\mathrm{cls}}\gets$ \Call{PathMap}{uniform, $\{F_q^{\mathrm{cls}}\}$}
\State $a^{\mathrm{path},\mathrm{s}},a^{\mathrm{path},\mathrm{e}}\gets$ \Call{PathMap}{cell-structured, $\{F_q^{\mathrm{s}},F_q^{\mathrm{e}}\}$}
\For{$\tau\in\{\mathrm{s},\mathrm{e}\}$}
  \For{$k=1,2,3$}
    \State $\ell_k\gets\max(1,\operatorname{round}(\rho_k|I_q|))$
    \State $H_k^\tau\gets$ the $\ell_k$ steps of $I_q$ next to boundary $\tau$
    \State $b^{(\tau,k)}\gets x$ with $H_k^\tau$ replaced by $m$
    \State $a^{(\tau,k)}\gets$ IG of $F_q^\tau$ from $b^{(\tau,k)}$ to $x$
  \EndFor
  \State $a^\tau\gets$ fuse $a^{\mathrm{path},\tau}$ and $a^{(\tau,1)},a^{(\tau,2)},a^{(\tau,3)}$ (Eq.~(5))
\EndFor
\State \Return $a^{\mathrm{cls}},a^{\mathrm{s}},a^{\mathrm{e}}$
\end{algorithmic}
\end{algorithm}

\section{Evaluation Details}
\label{sec:app-setup}

\subsection{Evaluation Protocol}
\label{sec:app-eval-protocol}
\paragraph{Features.}
Both detectors use the snippet features released with ActionFormer~\cite{zhang2022actionformer}: two-stream I3D features for THUMOS14, from 16-frame clips with a stride of 4 frames, and SlowFast features for EPIC-KITCHENS-100, from 32-frame clips with a stride of 16 frames.

\paragraph{Candidates.}
Final detections are first matched one-to-one to ground-truth instances of the same class at tIoU 0.5, greedily in decreasing score as in mAP evaluation, so that each instance is paired with the detection counted as its true positive rather than with a lower-scoring duplicate.
Matched detections with tIoU at least 0.7 form the candidate pool.
Candidates are selected from this pool deterministically by rotating across pyramid levels, with at most one candidate per video per round: one round for THUMOS14 and two for EPIC-KITCHENS-100.
For TriDet, the 199 THUMOS14 candidates come from 199 of the 212 test videos, and the 272 EPIC-KITCHENS-100 candidates from 137 of the 138 validation videos; the remaining videos have no matched detection with tIoU at least 0.7.
The videos are split into two halves with a fixed seed, and candidates are selected within each half; the development subset is one half, with 99 THUMOS14 candidates from 99 videos and 137 EPIC-KITCHENS-100 candidates from 69 videos.
Method comparisons are paired by candidate. 
Sampled orders, donor videos and replacement windows, and ranking use deterministic seeds.

\paragraph{Donor replacement.}
The perturbation tests replace removed steps with real features of another video rather than with the dataset mean $m$, following the real-point references of Hama \etal~\cite{hama2023deletion}; mean replacement sets many steps to the same vector, which can produce repeated, potentially out-of-distribution inputs~\cite{hooker2019benchmark,hama2023deletion}.
Donors are drawn from the other videos of the same split whose annotations exclude the explained class. A segment of a donor's features with the length of the input is taken at a random offset, or repeated cyclically if the donor is shorter, and a removed step $t$ takes step $t$ of this segment.

\paragraph{Class deletion and insertion.}
Let $F_x$ and $F_0$ be the class scores of the clean and fully replaced inputs, respectively.
All input steps are ranked by attribution; $F_{\mathrm{del}}(n)$ is the score after removing the $n$ highest-ranked steps from the clean input, and $F_{\mathrm{ins}}(n)$ the score after restoring them into the fully replaced input.
We evaluate the normalized deletion drop $(F_x-F_{\mathrm{del}}(n))/F_x$ and insertion gain $(F_{\mathrm{ins}}(n)-F_0)/F_x$ over $n=0,\ldots,L$, reporting their trapezoidal means.
The perturbation budget is $L=\min(N_{\mathrm{GT}},T)$, where $N_{\mathrm{GT}}$ is the duration of the matched ground-truth instance in input steps.
RISE~\cite{petsiuk2018rise} runs both curves to completion; detector evaluations instead scale the perturbation to the size of the explained object, since object sizes vary widely~\cite{jiang2023nagigos,zhao2024gradient}. Action lengths vary widely as well, so we stop at the length of the explained instance. RISE also uses the raw class probability with constant and blurred replacements; the donor replacement, the normalization by $F_x$, and the gain measured from $F_0$ are ours.

\begin{figure*}[t]
  \centering
  \includegraphics[width=\textwidth]{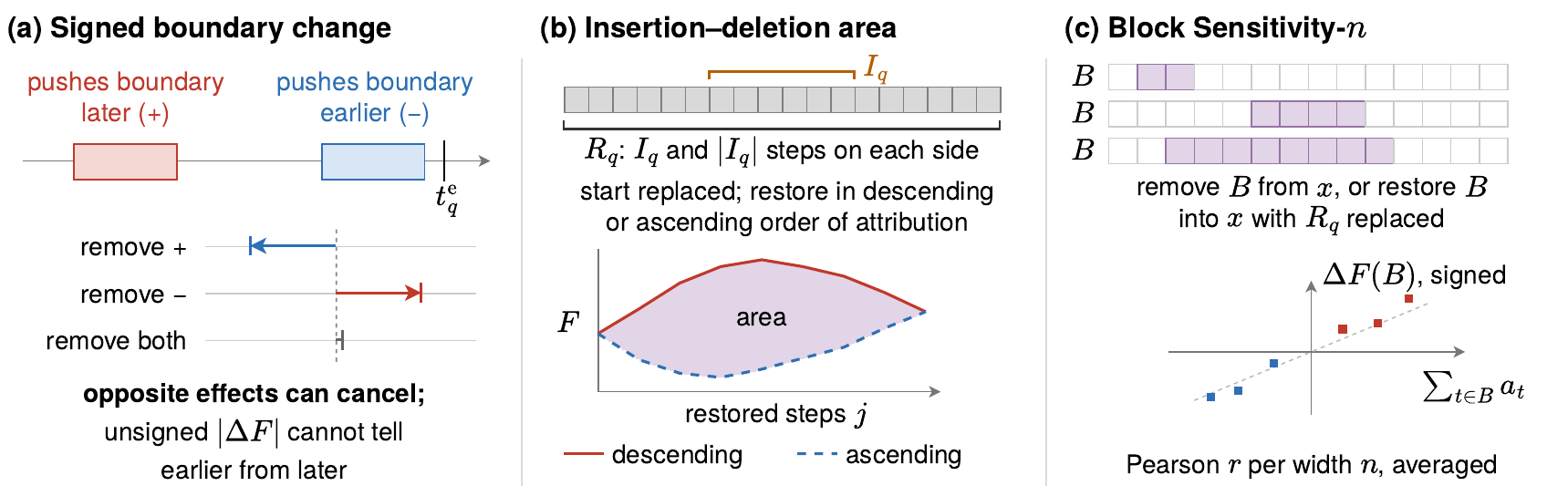}
  \caption{\textbf{Boundary faithfulness metrics} (schematic).
  (a) Removing evidence that pushes a boundary later moves it earlier, and vice versa, so an unsigned change cannot tell the two apart and opposite effects can cancel; we therefore use signed boundary changes.
  (b) Insertion--deletion area: starting with $R_q$ replaced, steps are restored in descending and in ascending order of attribution; the area between the two boundary curves, normalized by $|R_q|$ and the predicted duration, has zero expectation under random rankings (\cref{eq:app-id-area}).
  (c) Block Sensitivity-$n$: contiguous blocks of each width are removed from the clean input or restored into the input with $R_q$ replaced, and their summed attributions are correlated with the signed output changes.}
  \label{fig:app-eval-metrics}
\end{figure*}

\paragraph{Boundary insertion--deletion area.}
Let $R_q$ be the predicted interval $I_q$ extended by $|I_q|$ steps on both sides and clipped to the video (\cref{fig:app-eval-metrics}b).
Starting with all steps in $R_q$ replaced and the rest of the input clean, we restore them in descending and in ascending order of attribution; ascending restoration corresponds to deletion in reverse.
Let $F_{\mathrm{desc}}(j)$ and $F_{\mathrm{asc}}(j)$ denote the boundary predictions after restoring $j$ steps in each order, and let $D=F_q^{\mathrm{e}}(x)-F_q^{\mathrm{s}}(x)$ be the clean predicted duration.
The insertion--deletion area is
\begin{equation}
\mathcal{A}_{\mathrm{ID}}
=
\frac{1}{|R_q|}
\sum_{j=0}^{|R_q|}
\frac{F_{\mathrm{desc}}(j)-F_{\mathrm{asc}}(j)}{D}.
\label{eq:app-id-area}
\end{equation}
Following Hama \etal~\cite{hama2023deletion}, we use the signed area between the two curves; unlike the area under either curve alone, it has zero expectation under random attribution rankings, even when input features interact. The two normalizations are ours: dividing by $|R_q|$ makes the horizontal axis the fraction of restored steps, and dividing by $D$ measures boundary positions in units of the predicted duration, so that short and long candidates contribute on the same scale. Both are constant for a given candidate and therefore preserve the zero expectation.

\paragraph{Block Sensitivity-$n$.}
\Cref{fig:app-eval-metrics}c illustrates this metric. Blocks are taken at every position in $R_q$, and steps outside $R_q$ stay clean. Removal replaces $B$ in the clean input; restoration restores $B$ into the input with all of $R_q$ replaced and measures the change from that input.
For each candidate and width, the correlation is computed per donor and averaged over the donors; a correlation that is undefined because the attributions or the output changes are constant is skipped for that donor and counts as zero if it is undefined for all of them, and a candidate with no width of at least three blocks scores zero.

\paragraph{Aggregation.}
Each metric is computed per candidate, averaged over three donors, and then averaged over candidates; boundary metrics are computed for the start and the end, each with its own map, and averaged over the two.

\subsection{Attribution Implementations}
\label{sec:app-implementation}

\paragraph{Implementation settings.}
$m$ is the mean snippet feature over the training videos, each video weighted equally.
The receptive-field window half-widths are given in \cref{sec:app-path-approx}.

\paragraph{Compared methods.}
\Cref{tab:app-baselines} summarizes the implementation settings of the compared attribution methods.

\begin{table}[t]
\centering
\caption{\textbf{Compared attribution methods.}
Implementation settings for input-level attribution
of a fixed detector candidate.}
\label{tab:app-baselines}
\small
\setlength{\tabcolsep}{4pt}
\begin{tabularx}{\columnwidth}{@{}lX@{}}
\toprule
Method & Implementation settings \\
\midrule
IG~\cite{sundararajan2017axiomatic}
& Dataset-mean reference; straight-line path;
100 integration intervals. \\

Guided IG~\cite{kapishnikov2021guided}
& Dataset-mean reference; 25\% lowest-gradient
coordinates per step; maximum deviation 0.02
from the straight line; 100 steps. \\

Adaptive nodes~\cite{nakai2026adaptive}
& Adaptive integration-node placement on the
input straight-line path; 100 nodes. \\

Left to right
& Restore the context unit, then the window
steps one at a time from left to right. \\

Right to left
& Restore the context unit, then the window
steps one at a time from right to left. \\

TIMING~\cite{jang2025timing}
& 50 masked segments of 10--48 input steps;
all feature channels masked together. \\
\bottomrule
\end{tabularx}
\end{table}

For Adaptive IG-ODAM~\cite{nakai2026adaptive}, we apply only its adaptive integration-node placement, rather than reproducing the full object-detection attribution pipeline.
ODAM~\cite{zhao2023odam} and IG-ODAM~\cite{nakai2025igodam} operate on intermediate detector features rather than input snippets.
Adapting them to input-level attribution requires additional design choices; we therefore compare the adaptive integration-node strategy on the same input-level target as the other methods.

\paragraph{TIMING coverage.}
We run TIMING with its default segment settings: at each of the 100 integration nodes, 50 randomly placed segments of 10--48 steps are masked, i.e., keep their input values, and only the other steps move along the path. Steps covered at every node receive zero attribution, mostly in short inputs: on average, 36\% of the steps of an EPIC-KITCHENS-100 input (median length 417 steps) are never on the path, versus 7\% on THUMOS14 (median about 1,260 steps). With ActionFormer, whose input gradients vanish outside the receptive-field window, the class map is numerically zero for 33 of the 269 EPIC-KITCHENS-100 candidates, and at least one of the class, start, and end maps for 34, all shorter than 300 steps; TriDet's maps remain nonzero because its gradients reach the uncovered steps near the video ends.
On the candidates where TIMING perturbs at least 90\% of the steps (\cref{tab:app-timing-coverage}), our maps remain ahead on every class metric and in boundary insertion--deletion area and restoration Sensitivity-$n$ in all four settings.
TIMING is ahead in boundary removal Sensitivity-$n$ in three of them, consistent with Sec.~4 of the main paper: it attributes each step with most of the video kept at its input values, as in the removal test.

\begin{table}[t]
  \centering
  \footnotesize
  \setlength{\tabcolsep}{2.2pt}
  \begin{tabular}{@{}ll cccc ccc@{}}
    \toprule
    & & \multicolumn{4}{c}{Class} & \multicolumn{3}{c}{Boundary} \\
    \cmidrule(lr){3-6}\cmidrule(l){7-9}
    & & & & \multicolumn{2}{c}{Sens.-$n$} & Ins.-- & \multicolumn{2}{c}{Sens.-$n$} \\
    \cmidrule(lr){5-6}\cmidrule(l){8-9}
    & Map & Del. & Ins. & rem. & rest. & Del. & rem. & rest. \\
    \midrule
    \multicolumn{9}{@{}l}{\textbf{TriDet}} \\
    TH (176) & TIMING & 60.5 & 40.7 & 65.7 & 75.3 & 23.8 & \textbf{46.6} & 42.5 \\
             & Ours & \textbf{63.5} & \textbf{44.1} & \textbf{68.5} & \textbf{81.5} & \textbf{25.9} & 44.3 & \textbf{48.3} \\
    EP (132) & TIMING & 52.1 & 37.3 & 64.2 & 65.3 & 19.0 & \textbf{34.6} & 31.0 \\
             & Ours & \textbf{58.5} & \textbf{44.3} & \textbf{72.0} & \textbf{76.4} & \textbf{24.1} & 32.3 & \textbf{37.4} \\
    \midrule
    \multicolumn{9}{@{}l}{\textbf{ActionFormer}} \\
    TH (176) & TIMING & 63.1 & 39.4 & 68.6 & 78.6 & 22.8 & 48.2 & 41.8 \\
             & Ours & \textbf{65.2} & \textbf{41.2} & \textbf{70.8} & \textbf{84.5} & \textbf{25.3} & \textbf{49.4} & \textbf{46.4} \\
    EP (136) & TIMING & 50.1 & 36.3 & 60.4 & 61.0 & 13.1 & \textbf{26.2} & 16.0 \\
             & Ours & \textbf{55.2} & \textbf{42.2} & \textbf{67.5} & \textbf{70.6} & \textbf{16.9} & 25.0 & \textbf{18.0} \\
    \bottomrule
  \end{tabular}
  \caption{\textbf{TIMING on the candidates it covers} (candidates for which TIMING perturbs at least 90\% of the input steps, number in parentheses; donor replacement, $\times100$). Columns as in Tabs.~1 and 2 of the main paper; Ours: our class map and full boundary map, hybrid path. Bold: the better of the two.}
  \label{tab:app-timing-coverage}
\end{table}

\subsection{Runtime Measurement}
\label{sec:app-runtime}

Runtime is measured on one V100 for TriDet on THUMOS14 and on one A5500 for the other three settings.
The hybrid selects the faster method per candidate using GPU-specific measured costs.
Soft-path runtime is estimated from two sampled orders per candidate and scaled to $R=32$.
\section{Additional Results}
\label{sec:app-results}

\subsection{Temporal Redundancy and Order Averaging}
\label{sec:exp-redundancy}

We examine temporal interactions and order-dependent contributions that motivate order averaging, followed by ablations of the number of orders and ramp width.
\Cref{tab:exp-context} reports results on TriDet's development candidates with donor replacement.

\begin{table*}[!t]
  \centering
  \footnotesize
  \begin{minipage}[t]{0.43\textwidth}
    \centering
    \setlength{\tabcolsep}{3pt}
    \begin{tabular}[t]{@{}l ccccc@{}}
      \toprule
      \multicolumn{6}{@{}l}{\textbf{(a) Class-score drop, half of $I_q$ replaced}} \\
      & Contig. & Random & Even & & \\
      \midrule
      THUMOS14 & 65.3 & 55.2 & 49.5 & & \\
      EPIC-KITCHENS & 54.5 & 46.5 & 38.1 & & \\
      \midrule
      \multicolumn{6}{@{}l}{\textbf{(b) Joint block effect / sum of step effects}} \\
      Block width & 2 & 4 & 8 & & \\
      \midrule
      THUMOS14 & 1.56 & 1.69 & 1.76 & & \\
      EPIC-KITCHENS & 1.39 & 1.37 & 1.06 & & \\
      \midrule
      \multicolumn{6}{@{}l}{\textbf{(c) Per-order contribution of a step}} \\
      & \multicolumn{2}{c}{Neighbors} & \multicolumn{3}{c}{Across-order CV} \\
      \cmidrule(lr){2-3}\cmidrule(l){4-6}
      & Neither & Both & Class & Start & End \\
      \midrule
      THUMOS14 & 1.14 & 0.50 & 1.21 & 1.88 & 1.96 \\
      EPIC-KITCHENS & 1.22 & 0.34 & 1.06 & 2.54 & 2.62 \\
      \bottomrule
    \end{tabular}
  \end{minipage}\hfill
  \begin{minipage}[t]{0.54\textwidth}
    \centering
    \setlength{\tabcolsep}{3pt}
    \begin{tabular}[t]{@{}l cccccc@{}}
      \toprule
      \multicolumn{7}{@{}l}{\textbf{(d) Orders: step deletion minus IG}} \\
      Orders $R$ & 1 & 2$^{\ddagger}$ & 4 & 8 & 16 & 32 \\
      \midrule
      THUMOS14 & $-8.8$ & $-4.0$ & $-1.8$ & $-0.1$ & 1.0 & 1.4 \\
      EPIC-KITCHENS & $-4.6$ & 0.9 & 3.3 & 4.4 & 5.4 & 5.8 \\
      \midrule
      \multicolumn{7}{@{}l}{\textbf{(e) Ramp width $w$}} \\
      $w$ & 0$^{\ast}$ & 0.05 & 0.1 & 0.3 & 0.5 & 1 (IG) \\
      \midrule
      \multicolumn{7}{@{}l}{\emph{THUMOS14}} \\
      Del. & 64.75 & 64.89 & 65.07 & 65.38 & 65.05 & 63.36 \\
      Ins. & 42.23 & 42.23 & 42.14 & 41.79 & 40.97 & 38.67 \\
      CV & 1.30 & 1.28 & 1.21 & 0.97 & 0.69 & n/a \\
      \multicolumn{7}{@{}l}{\emph{EPIC-KITCHENS-100}} \\
      Del. & 56.09 & 56.09 & 56.04 & 55.58 & 54.32 & 50.30 \\
      Ins. & 43.43 & 43.27 & 43.28 & 42.63 & 41.46 & 36.93 \\
      CV & 1.11 & 1.11 & 1.06 & 0.90 & 0.69 & n/a \\
      \bottomrule
    \end{tabular}
  \end{minipage}
  \caption{\textbf{Temporal redundancy and order averaging} (TriDet, development candidates, donor replacement, class score unless stated). (a) Class-score drop ($\times100$) when half of the clean predicted interval $I_q$ is replaced as one contiguous span, at random steps, or at evenly spaced steps. (b) Median ratio of the class-score change after replacing a contiguous block to the sum of the changes from replacing its steps one at a time. (c) Contribution of a step inside $I_q$ along windowed soft orders when neither or both immediate neighbors are present, in units of the mean absolute attribution within $I_q$; CV is the coefficient of variation of a step's contribution across orders (its standard deviation divided by its absolute mean), with the median taken over the steps in $I_q$ and then over candidates. (d) Step-deletion difference ($\times100$) to straight-line IG when averaging the first $R$ sampled orders, with exact forward differences; $^{\ddagger}$an order and its reverse. (e) Step deletion and insertion ($\times100$) and across-order CV with 32 fixed orders per width; $^{\ast}$forward differences.}
  \label{tab:exp-context}
\end{table*}

\paragraph{Temporal redundancy.}
Replacing one contiguous half of $I_q$ lowers the class score more than replacing the same number of randomly chosen or evenly spaced steps (\cref{tab:exp-context}a).
Similarly, replacing a contiguous block of two to eight steps produces a score change 1.06--1.76 times the sum of the changes from replacing its steps individually (\cref{tab:exp-context}b).
These results indicate that single-step perturbations can understate the joint effect of contiguous evidence, consistent with local redundancy, although other nonlinear effects may also contribute.

\paragraph{Order-dependent contributions.}
Along sampled windowed orders, a step contributes 1.14 and 1.22 times the mean absolute attribution within $I_q$ when neither immediate neighbor has been restored, compared with 0.50 and 0.34 when both have (\cref{tab:exp-context}c).
Across orders, the coefficient of variation (CV) of per-step contributions is 1.21 and 1.06 for the class score, but 1.88--2.62 for the boundaries; ActionFormer shows a similar difference (1.10 and 1.12 versus 1.68--3.37).
These variations support averaging contributions across restoration contexts rather than relying on a single path, particularly for boundary attribution.

\paragraph{Number of orders and ramp width.}
A single restoration order performs worse than straight-line IG in step deletion, whereas averaging more orders improves faithfulness (\cref{tab:exp-context}d).
The improvement over IG becomes positive at $R=16$ on THUMOS14 and $R=2$ on EPIC-KITCHENS-100, reaching 1.4 and 5.8 points at $R=32$, with diminishing gains as $R$ increases. 
For the ramp width $w$, the adopted $w=0.1$ remains within 0.32 points of the best tested width in step deletion and insertion (\cref{tab:exp-context}e).
Wide ramps ($w\geq0.5$) lower faithfulness toward straight-line IG, and the across-order CV decreases as $w$ grows.
We therefore use the soft path where it is faster than exact forward differences, rather than for variance reduction.

\subsection{Class Faithfulness of Cell-Structured Orders}
\label{sec:app-class-cells}

\Cref{tab:app-class-cells} evaluates cell-structured orders for class attribution, in addition to the windowed random orders used by our class map.
Adding cell-structured orders to the windowed orders lowers deletion, insertion, and removal Sensitivity-$n$ in all four settings, by 0.2--2.6 points, and lowers restoration Sensitivity-$n$ on EPIC-KITCHENS-100, while changing it by at most 0.3 on THUMOS14.
This is consistent with the smaller across-order CV of class contributions than of boundary contributions (1.06--1.21 versus 1.68--3.37, \cref{sec:exp-redundancy}): class contributions depend less on the restoration order, which cell grouping constrains.
Unlike for boundaries, cell grouping therefore does not benefit class attribution, and our class map uses windowed orders without cells.

\begin{table}[t]
  \centering
  \footnotesize
  \setlength{\tabcolsep}{2.6pt}
  \begin{tabular}{@{}l cccc cccc@{}}
    \toprule
    & \multicolumn{4}{c}{THUMOS14} & \multicolumn{4}{c}{EPIC-KITCHENS} \\
    \cmidrule(lr){2-5}\cmidrule(l){6-9}
    & & & \multicolumn{2}{c}{Sens.-$n$\,$\uparrow$} & & & \multicolumn{2}{c}{Sens.-$n$\,$\uparrow$} \\
    \cmidrule(lr){4-5}\cmidrule(l){8-9}
    Orders & Del.$\uparrow$ & Ins.$\uparrow$ & rem. & rest. & Del.$\uparrow$ & Ins.$\uparrow$ & rem. & rest. \\
    \midrule
    \multicolumn{9}{@{}l}{\textbf{TriDet}} \\
    Unstructured & 63.0 & 43.2 & 68.3 & 81.4 & 56.3 & 41.3 & 71.1 & 73.7 \\
    \quad + window & 63.3 & 43.8 & 68.2 & 81.2 & 56.1 & 41.3 & 70.9 & 73.5 \\
    \quad + cells & 62.8 & 43.2 & 67.6 & 81.5 & 55.1 & 38.8 & 69.2 & 71.9 \\
    \midrule
    \multicolumn{9}{@{}l}{\textbf{ActionFormer}} \\
    Unstructured & 64.5 & 41.0 & 70.6 & 84.8 & 52.9 & 40.8 & 68.7 & 71.7 \\
    \quad + window & 64.6 & 41.1 & 70.6 & 84.6 & 52.9 & 40.9 & 68.8 & 71.6 \\
    \quad + cells & 64.1 & 40.9 & 70.4 & 84.9 & 52.2 & 38.9 & 67.7 & 70.9 \\
    \bottomrule
  \end{tabular}
  \caption{\textbf{Class-score faithfulness with cell-structured
    orders} (donor replacement, soft path, $\times100$).
    Columns as in Tab.~1 of the main paper.
    + window: ordering restricted to the receptive-field window
    (our class map);
    + cells: additional cell-structured ordering, as used
    for boundary attribution.}
  \label{tab:app-class-cells}
\end{table}

\subsection{Boundary Maps per Pyramid Level}
\label{sec:app-levels}

\Cref{tab:app-levels} breaks down removal Sensitivity-$n$ gains by pyramid level on EPIC-KITCHENS-100, comparing exact forward differences with soft paths and cell-structured with unstructured window orders.
Exact differences and soft paths differ by at most 0.3 points at levels 0--1, whereas exact differences
are 3.5--9.6 points higher at levels 3--5, concentrating the integration-method gap at
coarser levels, possibly because more ordered units transition simultaneously along the soft path as the window grows (Eq.~(3) of the main paper).
Cell grouping changes removal Sensitivity-$n$ by at most 1.6 points at levels 0--1, where cells contain one or two input steps, but improves it over unstructured window orders by 5.9--15.3 points at levels 3--5.
On THUMOS14, the cell gain with soft paths follows the same pattern: at most 0.4 points at levels 0--1 and 1.3--6.9 points at levels 3--5.
The overall gains from cell grouping are comparable under soft paths and exact differences
(4.1 versus 4.6 for TriDet and 4.1 versus 3.8 for ActionFormer), indicating that its benefit is not specific to the soft-path approximation.

\begin{table}[!t]
  \centering
  \footnotesize
  \setlength{\tabcolsep}{1.8pt}
  \begin{tabular}{@{}c cccc cccc@{}}
    \toprule
    & \multicolumn{4}{c}{TriDet}
    & \multicolumn{4}{c}{ActionFormer} \\
    \cmidrule(lr){2-5}\cmidrule(l){6-9}
    & & & \multicolumn{2}{c}{Cells$-$window}
    & & & \multicolumn{2}{c}{Cells$-$window} \\
    \cmidrule(lr){4-5}\cmidrule(l){8-9}
    Level & $n$ & Exact$-$soft & soft & exact
          & $n$ & Exact$-$soft & soft & exact \\
    \midrule
    0 & 56 & $-$0.3 & 0.3 & 0.5
      & 57 & $-$0.2 & $-$0.9 & $-$0.6 \\
    1 & 57 & 0.1 & $-$0.5 & $-$0.5
      & 57 & 0.0 & 1.6 & 1.5 \\
    2 & 54 & 1.2 & 1.0 & 2.4
      & 54 & 2.9 & 2.9 & 4.3 \\
    3 & 53 & 3.5 & 6.0 & 8.5
      & 53 & 5.5 & 7.5 & 8.5 \\
    4 & 44 & 8.0 & 14.3 & 13.6
      & 42 & 7.3 & 10.3 & 5.9 \\
    5 & 8 & 9.6 & 15.3 & 10.0
      & 10 & 7.8 & 9.8 & 6.3 \\
    \midrule
    All & 272 & 2.5 & 4.1 & 4.6
        & 273 & 3.0 & 4.1 & 3.8 \\
    \bottomrule
  \end{tabular}
  \caption{\textbf{Boundary faithfulness gains by pyramid level}
  on EPIC-KITCHENS-100 (removal Sensitivity-$n$, donor replacement, $\times100$, averaged over start and end).
  Each column is the difference between two maps that differ in one design choice.
  Exact$-$soft: exact forward differences minus soft paths, both with unstructured window orders.
  Cells$-$window: cell-structured minus unstructured window orders, both with soft paths or both with exact differences.
  $n$: candidates at the level; a level-$l$ cell spans about $2^l$ input steps.}
  \label{tab:app-levels}
\end{table}

\subsection{Block Sensitivity-\texorpdfstring{$n$}{n} by Width}
\label{sec:app-widths}

\begin{table*}[!t]
  \centering
  \footnotesize
  \setlength{\tabcolsep}{5pt}
  \begin{tabular}{@{}l ccccccc@{}}
    \toprule
    & \multicolumn{7}{c}{Block width $|B|$ (input steps)} \\
    \cmidrule(l){2-8}
    Map & 1 & 2 & 4 & 8 & 16 & 32 & $|I_q|$ \\
    \midrule
    \multicolumn{8}{@{}l}{\emph{THUMOS14, removal}} \\
    IG & 33.9 & 33.2 & 33.0 & 35.0 & 33.5 & 29.5 & 43.2 \\
    Unstructured & 34.6 & 36.0 & 36.4 & 37.8 & 36.0 & 32.1 & 47.5 \\
    + cells (ours) & 34.9 & 36.8 & 38.9 & 41.4 & 40.2 & 37.8 & 51.4 \\
    + references (ours) & 38.4 & 41.0 & 43.5 & 45.6 & 43.6 & 41.2 & 53.4 \\
    \midrule
    \multicolumn{8}{@{}l}{\emph{THUMOS14, restoration}} \\
    IG & 34.5 & 41.1 & 45.7 & 47.9 & 47.2 & 37.7 & 42.4 \\
    Unstructured & 38.5 & 47.4 & 52.7 & 53.8 & 51.4 & 42.0 & 46.6 \\
    + cells (ours) & 36.6 & 46.2 & 53.6 & 56.1 & 54.8 & 47.0 & 51.0 \\
    + references (ours) & 31.5 & 40.9 & 49.1 & 53.1 & 54.3 & 49.9 & 53.7 \\
    \midrule
    \multicolumn{8}{@{}l}{\emph{EPIC-KITCHENS-100, removal}} \\
    IG & 18.1 & 10.4 & 8.3 & 5.0 & 3.4 & $-$0.5 & 15.6 \\
    Unstructured & 19.0 & 13.0 & 9.8 & 6.4 & 5.0 & 0.6 & 18.8 \\
    + cells (ours) & 19.8 & 15.9 & 15.0 & 13.0 & 11.5 & 7.2 & 22.0 \\
    + references (ours) & 30.3 & 30.4 & 31.8 & 33.9 & 31.0 & 24.0 & 37.3 \\
    \midrule
    \multicolumn{8}{@{}l}{\emph{EPIC-KITCHENS-100, restoration}} \\
    IG & 23.7 & 29.2 & 32.9 & 29.7 & 15.6 & 2.2 & 17.2 \\
    Unstructured & 30.3 & 38.9 & 42.4 & 34.6 & 17.1 & 3.5 & 21.4 \\
    + cells (ours) & 28.6 & 36.7 & 41.8 & 37.7 & 22.2 & 8.7 & 25.5 \\
    + references (ours) & 26.1 & 33.9 & 41.9 & 43.7 & 35.7 & 25.4 & 41.5 \\
    \midrule
    Candidates, THUMOS14 & 199 & 199 & 199 & 199 & 183 & 149 & 186 \\
    Candidates, EPIC-KITCHENS-100 & 272 & 272 & 270 & 226 & 196 & 132 & 223 \\
    \bottomrule
  \end{tabular}
  \caption{\textbf{Signed block Sensitivity-$n$ of the boundary maps by width} (TriDet, soft path, donor replacement, Pearson correlation $\times100$, averaged over start and end). Widths 1--32 are the cell sizes of TriDet's pyramid levels 0--5; $|I_q|$ is the predicted interval length and has its own column only when it is not a power of two. A width counts for a candidate only if at least three blocks of that width fit into $R_q$, so the columns cover different numbers of candidates and do not average to the block Sensitivity-$n$ of Tab.~2 of the main paper, which first averages each candidate over its own widths.}
  \label{tab:app-widths}
\end{table*}

\begin{table}[!t]
  \centering
  \footnotesize
  \setlength{\tabcolsep}{3pt}
  \begin{tabular}{@{}ll ccc cc@{}}
    \toprule
    & & \multicolumn{3}{c}{$|B|$ vs.\ cell width $s_l$} & \multicolumn{2}{c}{$|B|=s_l$} \\
    \cmidrule(lr){3-5}\cmidrule(l){6-7}
    Dataset & Sensitivity-$n$ & $<s_l$ & $=s_l$ & $>s_l$ & One cell & Other \\
    \midrule
    \multicolumn{7}{@{}l}{\textbf{TriDet}} \\
    THUMOS14 & removal & 0.9 & 3.9 & 4.8 & 4.5 & 4.5 \\
             & restoration & $-0.7$ & 3.2 & 4.1 & 3.9 & 3.5 \\
    EPIC-KITCHENS & removal & 3.6 & 7.5 & 4.9 & 18.3 & 6.4 \\
                  & restoration & $-1.6$ & 4.0 & 4.5 & 9.3 & 4.1 \\
    \midrule
    \multicolumn{7}{@{}l}{\textbf{ActionFormer}} \\
    THUMOS14 & removal & $-0.1$ & 2.1 & 2.7 & 2.3 & 2.6 \\
             & restoration & $-0.5$ & 1.9 & 3.0 & 1.7 & 2.2 \\
    EPIC-KITCHENS & removal & 6.0 & 8.0 & 3.2 & 18.7 & 5.8 \\
                  & restoration & $-2.4$ & 2.7 & 3.4 & 9.4 & 2.6 \\
    \bottomrule
  \end{tabular}
  \caption{\textbf{Gain of cell-structured over windowed ordering by block width relative to the cell} (boundary
  block Sensitivity-$n$, soft path, donor replacement, difference of Pearson correlations $\times100$, averaged over start and end). Left:
  blocks narrower than, equal to, or wider than the candidate's cell width $s_l=2^l$; per candidate the mean over its
  widths in each group, averaged over the candidates at levels 1--4, where all three groups exist (TriDet: 158
  THUMOS14 and 208 EPIC-KITCHENS-100; ActionFormer: 164 and 206). Right: blocks of width $s_l$ that cover exactly one
  level-$l$ cell of Eq.~(4) of the main paper versus blocks at the other offsets, levels 1--5 (TriDet: 178 and 216;
  ActionFormer: 185 and 214).}
  \label{tab:app-widths-cell}
\end{table}

\paragraph{Evaluation block width.}
\Cref{tab:app-widths} reports boundary block Sensitivity-$n$ separately for different block widths. Under removal, the ordering IG $<$ unstructured $<$ cell-structured $<$
boundary references holds at every width.
Under restoration, the comparison depends on scale: unstructured averaging performs better at smaller widths, while cell grouping and boundary references become more effective for wider blocks.
In particular, boundary references achieve the highest correlations at the widest widths but underperform IG for one- and two-step blocks on THUMOS14.

\paragraph{Block width relative to pyramid cells.}
\Cref{tab:app-widths-cell} examines the gain from cell-structured ordering relative to the candidate's cell width $s_l=2^l$.
Under restoration, the gain is negative for blocks narrower than a cell and positive at or above the cell width across both detectors and datasets.
Under removal, the gain peaks at the cell width on EPIC-KITCHENS-100 and keeps growing for wider blocks on THUMOS14.
For blocks matching the cell width, alignment with a single pyramid cell yields larger gains on EPIC-KITCHENS-100 but not consistently on THUMOS14.
Thus, the results suggest a scale-dependent benefit of cell grouping, but no consistent benefit of alignment with a single cell across datasets.

\subsection{Boundary Reference Design}
\label{sec:app-references}
We compare boundary-anchored references with randomly positioned spans of the same widths and examine their combination with cell-structured order averaging (\cref{tab:exp-references}).

\paragraph{Reference placement.}
Under straight-line IG, boundary-anchored references outperform randomly positioned spans of the same widths across all boundary metrics on both datasets.
The gains in removal and restoration Sensitivity-$n$ are 2.2 and 1.9 points on THUMOS14, and 2.2 and 3.2 points on EPIC-KITCHENS-100, respectively.
This indicates that reference placement matters beyond the number and widths of the masked spans.

\paragraph{Reference and path design.}
Boundary-anchored references improve removal Sensitivity-$n$ over global IG even with a
straight-line path, from 35.6 to 41.7 on THUMOS14 and from 13.6 to 27.4 on EPIC-KITCHENS-100.
Conversely, cell-structured order averaging without local references improves the same metric to 40.9 and 19.2, respectively.
These comparisons show that the two components can each improve boundary faithfulness without requiring the other.

\paragraph{Fusion weight.}
Fusing boundary-anchored references with the cell-structured map at $\lambda=1$ further improves removal Sensitivity-$n$ on both datasets, but reduces restoration Sensitivity-$n$ on THUMOS14.
Increasing the global-path weight to $\lambda=2$ moderates this restoration loss (47.6 to 48.4), while reducing removal Sensitivity-$n$ on both datasets (44.3 to 43.7 on THUMOS14 and 33.0 to 29.6 on EPIC-KITCHENS-100).
We adopt $\lambda=1$ as the default, while acknowledging this trade-off between removal
and restoration faithfulness.

\begin{table}[!t]
  \centering
  \footnotesize
  \setlength{\tabcolsep}{1.3pt}
  \begin{tabular}{@{}l ccc ccc@{}}
    \toprule
    & \multicolumn{3}{c}{THUMOS14} & \multicolumn{3}{c}{EPIC-KITCHENS} \\
    \cmidrule(lr){2-4}\cmidrule(l){5-7}
    & & \multicolumn{2}{c}{Sens.-$n$\,$\uparrow$} & & \multicolumn{2}{c}{Sens.-$n$\,$\uparrow$} \\
    \cmidrule(lr){3-4}\cmidrule(l){6-7}
    Reference & Ins.--Del.$\uparrow$ & rem. & rest. & Ins.--Del.$\uparrow$ & rem. & rest. \\
    \midrule
    \multicolumn{7}{@{}l}{\emph{Straight line}} \\
    Dataset mean (IG) & 22.2 & 35.6 & 43.2 & 15.0 & 13.6 & 25.6 \\
    Random spans in $I_q$ & 22.7 & 39.5 & 41.6 & 19.0 & 25.2 & 28.8 \\
    Boundary spans & 23.9 & 41.7 & 43.5 & 19.4 & 27.4 & 32.0 \\
    \midrule
    \multicolumn{7}{@{}l}{\emph{Cell-structured random orders}} \\
    Dataset mean & 25.5 & 40.9 & 49.7 & 19.9 & 19.2 & 32.1 \\
    Boundary spans, $\lambda=1$ (ours) & 26.0 & 44.3 & 47.6 & 23.4 & 33.0 & 36.3 \\
    Boundary spans, $\lambda=2$ & 25.9 & 43.7 & 48.4 & 22.6 & 29.6 & 35.8 \\
    \bottomrule
  \end{tabular}
  \caption{\textbf{Reference design for the boundaries} (TriDet, soft path, donor replacement, $\times100$, averaged over start and end).
  Span maps are fused with the path map by Eq.~(5) of the main paper; random spans have the same widths at random positions inside the predicted interval.
  }
  \label{tab:exp-references}
\end{table}

\section{Analysis Details}
\label{sec:app-analysis}

\subsection{Cases and Maps}
\label{sec:app-counts}

\paragraph{Cases.}
For TH, EP, and AF, we use the candidates from Sec.~4 of the main paper.
For IV, the ground-truth instance of each TH candidate is matched to IV's final detections by the same rule (\cref{sec:app-eval-protocol}), yielding 159 candidates.
TH is the reference setting: EP changes the dataset, AF the detector, and IV the input features, so each factor is varied once relative to TH; we therefore do not add ActionFormer on EPIC-KITCHENS-100.
Both class and boundary maps use the soft path.

Unless otherwise stated, read-outs use cases whose predicted interval $I$ contains at least three input steps (199, 252, 159, and 198 cases in TH, EP, IV, and AF).
The profiles in Fig.~3a, b of the main paper use all cases.
Each case contributes its start and end boundaries, referred to as case-boundaries.
Confidence intervals are computed by resampling videos rather than individual case-boundaries; in particular, the 272 EP cases come from 137 videos.

\begin{table}[t]
  \centering
  \footnotesize
  \setlength{\tabcolsep}{3pt}
  \begin{tabular}{@{}lrrrr@{}}
    \toprule
    & TH & EP & IV & AF \\
    \midrule
    Strongest expanding step in the boundary's cell & & & & \\
    \quad or the next cell inside (\%) & 59 & 80 & 73 & 68 \\
    Class positive at the boundary's cell (\%) & 87 & 85 & 78 & 87 \\
    \quad and one cell beyond it (\%) & 71 & 66 & 49 & 67 \\
    \midrule
    \multicolumn{5}{@{}l}{\emph{The $k$ steps just inside the boundary}} \\
    Share of the near half's steps (\%) & 33 & 35 & 33 & 33 \\
    Share of its expanding support (\%) & 54 & 54 & 38 & 39 \\
    Class share / even share & 0.47 & 0.62 & 0.48 & 0.51 \\
    \bottomrule
  \end{tabular}
  \caption{\textbf{Map read-outs at the boundary}. Case-boundaries of cases with $|I|\geq3$; medians for the shares.
  Strongest expanding step: the step of largest expanding attribution of the fused boundary map within $|I|/2$ of the boundary, located in cells of the candidate's pyramid level; boundaries without positive expanding attribution there are skipped.
  Class attribution positive at a cell: the first cell, moving outward from the candidate's anchor cell within the receptive-field window, whose class-attribution sum is non-positive lies beyond it.
  Near half: the half of $I$ next to the boundary; expanding support: positive expanding attribution of the boundary map without references, for the boundaries whose near half has any.
  Class share: the $k$ steps' share of the box's class attribution, divided by $k/|I|$.}
  \label{tab:app-readouts}
\end{table}

\paragraph{Map read-outs.}
The strongest expanding step lies in the boundary's own cell or the next cell inside for 59--80\% of the evaluated boundaries (\cref{tab:app-readouts}).
Class attribution remains positive at the boundary's cell for 78--87\% of boundaries and one cell beyond for 49--71\%.
The $k$ steps just inside a boundary comprise about one-third of the near half and account for 38--54\% of its positive expanding support, but only 0.47--0.62 times an even share of the class attribution.

\begin{table*}[t]
  \centering
  \footnotesize
  \setlength{\tabcolsep}{4pt}
  \begin{tabular}{@{}llcccc@{}}
    \toprule
    & Boundary map & TH & EP & IV & AF \\
    \midrule
    Inside & Without references & $-$0.16 [$-$0.26, $-$0.06] & $-$0.11 [$-$0.20, $-$0.01] & $-$0.14 [$-$0.24, $-$0.02] & 0.01 [$-$0.10, 0.11] \\
     & \quad partial, $k/|I|$ & $-$0.11 [$-$0.21, $-$0.01] & $-$0.15 [$-$0.25, $-$0.05] & $-$0.10 [$-$0.20, 0.01] & 0.06 [$-$0.04, 0.15] \\
     & \quad partial, $k/|I|$ and level & $-$0.11 [$-$0.21, $-$0.01] & $-$0.17 [$-$0.27, $-$0.07] & $-$0.09 [$-$0.19, 0.01] & 0.06 [$-$0.03, 0.16] \\
     & Fused & $-$0.07 [$-$0.18, 0.04] & $-$0.17 [$-$0.25, $-$0.07] & 0.06 [$-$0.06, 0.19] & 0.01 [$-$0.11, 0.12] \\
    \midrule
    Outside & Without references & $-$0.07 [$-$0.17, 0.04] & $-$0.02 [$-$0.12, 0.07] & 0.18 [0.07, 0.30] & 0.12 [0.02, 0.22] \\
     & \quad partial, $k/|I|$ & $-$0.09 [$-$0.20, 0.02] & $-$0.06 [$-$0.16, 0.03] & 0.18 [0.07, 0.29] & 0.09 [$-$0.02, 0.19] \\
     & \quad partial, $k/|I|$ and level & $-$0.09 [$-$0.20, 0.02] & $-$0.07 [$-$0.16, 0.02] & 0.19 [0.07, 0.30] & 0.08 [$-$0.03, 0.19] \\
    \bottomrule
  \end{tabular}
  \caption{\textbf{Class and boundary maps on the same replaced steps} (background swap). Spearman correlation between the class map's and the boundary map's expectation across case-boundaries, with its 95\% video-bootstrap interval; partial: controlling for the replaced share $k/|I|$, or for it and the pyramid level. Fused: the boundary map with references; outside the box, where its references do not act, it equals the map without references. Case-boundaries: 398, 504, 318, and 396 inside and 393, 499, 311, and 391 outside.}
  \label{tab:app-map-rank}
\end{table*}

\begin{table*}[t]
  \centering
  \footnotesize
  \setlength{\tabcolsep}{2pt}
  \begin{tabular}{@{}lrrrrrrrr@{}}
    \toprule
    & \multicolumn{4}{c}{Inside the box}
    & \multicolumn{4}{c}{Outside the box} \\
    \cmidrule(lr){2-5}\cmidrule(l){6-9}
    & TH & EP & IV & AF
    & TH & EP & IV & AF \\
    \midrule
    \multicolumn{9}{@{}l}{\emph{Boundary: outward shift per replaced step}} \\
    Measured & $-$.46 & $-$.22 & $-$.61 & $-$.46 & $-$.18 & $-$.05 & $-$.20 & $-$.16 \\
    Expected & $-$.13 & $-$.05 & $-$.45 & $-$.28 & $-$.07 & +.13 & $-$.24 & $-$.04 \\
    \midrule
    \multicolumn{9}{@{}l}{\emph{Class score: relative change (\%)}} \\
    Measured & $-$8.6 & $-$10.2 & $-$8.8 & $-$7.9 & $-$3.1 & $-$5.6 & $-$1.9 & $-$2.5 \\
    Expected & $-$7.5 & $-$8.6 & $-$7.8 & $-$7.9 & $-$3.2 & $-$3.8 & $-$1.1 & $-$2.7 \\
    \midrule
    \multicolumn{9}{@{}l}{\emph{Spearman $\rho$ across case-boundaries}} \\
    Class map, score & .68 & .68 & .75 & .66 & .60 & .46 & .71 & .71 \\
    Boundary map, shift & .46 & .25 & .39 & .52 & .52 & .21 & .53 & .52 \\
    Class map, shift & $-$.16 & $-$.27 & $-$.08 & $-$.10 & $-$.02 & .07 & .07 & .04 \\
    Score, shift & $-$.07 & $-$.30 & $-$.01 & $-$.06 & .08 & .03 & .07 & .06 \\
    $k/|I|$, score & $-$.27 & $-$.52 & $-$.20 & $-$.23 & $-$.15 & $-$.27 & .01 & $-$.12 \\
    \bottomrule
  \end{tabular}
  \caption{\textbf{Background swap at the boundary.} The $k$ steps just inside or just outside a predicted boundary are replaced with background. Medians over case-boundaries, start and end pooled.
  Top: outward shift of the boundary per replaced step (inside, negative values are retractions); middle: relative change of the candidate's class score.
  Expected: minus the attribution of the replaced steps, for the boundary from the map without boundary-anchored references (the open markers of Fig.~3c, d of the main paper).
  Bottom: Spearman correlation of a map's expected change, the measured class-score change (score), or the replaced share $k/|I|$, with the measured class-score change (score) or boundary shift (shift).
  TH: TriDet on THUMOS14 (I3D); EP: TriDet on EPIC-KITCHENS-100 (SlowFast); IV: TriDet on THUMOS14 (InternVideo2); AF: ActionFormer on THUMOS14 (I3D).}
  \label{tab:analysis-swap}
\end{table*}

\begin{table*}[t]
  \centering
  \footnotesize
  \setlength{\tabcolsep}{4pt}
  \begin{tabular}{@{}llcccc@{}}
    \toprule
    & Class-score change & TH & EP & IV & AF \\
    \midrule
    Inside & $>-2\%$ & $-$0.46 [$-$0.52, $-$0.36] (104) & $-$0.33 [$-$0.43, $-$0.22] (117) & $-$0.67 [$-$0.76, $-$0.51] (79) & $-$0.44 [$-$0.56, $-$0.39] (94) \\
     & $-2$ to $-5\%$ & $-$0.48 [$-$0.56, $-$0.34] (50) & $-$0.31 [$-$0.47, $-$0.15] (53) & $-$0.55 [$-$0.68, $-$0.37] (32) & $-$0.46 [$-$0.56, $-$0.41] (48) \\
     & $-5$ to $-10\%$ & $-$0.48 [$-$0.59, $-$0.37] (60) & $-$0.32 [$-$0.40, $-$0.21] (80) & $-$0.54 [$-$0.75, $-$0.46] (55) & $-$0.43 [$-$0.54, $-$0.34] (86) \\
     & $\leq-10\%$ & $-$0.40 [$-$0.52, $-$0.30] (184) & $-$0.09 [$-$0.20, $-$0.03] (254) & $-$0.67 [$-$0.80, $-$0.55] (152) & $-$0.49 [$-$0.57, $-$0.41] (168) \\
    \midrule
    Outside & $>-2\%$ & $-$0.15 [$-$0.19, $-$0.10] (161) & $-$0.04 [$-$0.09, 0.01] (158) & $-$0.18 [$-$0.22, $-$0.14] (162) & $-$0.14 [$-$0.20, $-$0.09] (185) \\
     & $-2$ to $-5\%$ & $-$0.15 [$-$0.19, $-$0.12] (90) & $-$0.00 [$-$0.10, 0.05] (77) & $-$0.19 [$-$0.25, $-$0.09] (54) & $-$0.16 [$-$0.24, $-$0.09] (73) \\
     & $-5$ to $-10\%$ & $-$0.20 [$-$0.31, $-$0.10] (67) & $-$0.04 [$-$0.10, 0.05] (96) & $-$0.21 [$-$0.32, $-$0.14] (47) & $-$0.17 [$-$0.28, $-$0.08] (63) \\
     & $\leq-10\%$ & $-$0.27 [$-$0.33, $-$0.19] (75) & $-$0.06 [$-$0.09, $-$0.02] (168) & $-$0.31 [$-$0.46, $-$0.21] (48) & $-$0.21 [$-$0.30, $-$0.14] (70) \\
    \bottomrule
  \end{tabular}
  \caption{\textbf{Boundary shift by class-score change} (background swap; Fig.~3e, f of the main paper). Median outward shift of the boundary per replaced step, with its 95\% video-bootstrap interval and the number of case-boundaries; the groups are the relative change of the candidate's class score (the first includes rises).}
  \label{tab:app-bins}
\end{table*}

\subsection{Start Boundaries}
\label{sec:app-evidence-start}
\begin{figure}[!t]
  \centering
  \includegraphics[width=\columnwidth]{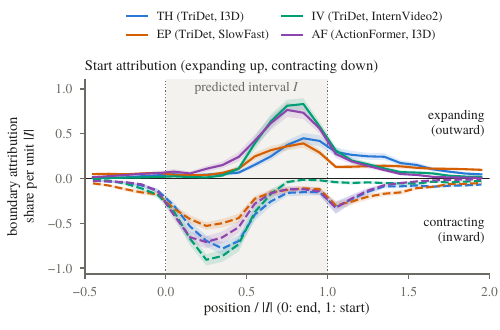}
  \caption{\textbf{Start boundary.} As Fig.~3b of the main paper, for the start, with the axis mirrored so that outward again points right (0: end; 1: start); positions above 1 cover up to $|I|$ before the start. The class profile for the start is Fig.~3a of the main paper read from right to left. Means with 95\% video-bootstrap intervals.}
  \label{fig:app-evidence-start}
\end{figure}

\Cref{fig:app-evidence-start} shows the profile of Fig.~3b of the main paper for the start boundary. For each boundary separately, in all four settings and with and without the boundary-anchored references, the median expanding share is larger in the half of $I$ next to the explained boundary than in the opposite half, and the median contracting share is larger in the opposite half, with 95\% video-bootstrap intervals excluding zero.

\subsection{Class and Boundary on the Same Steps}
\label{sec:app-same-steps}

We compare the class and boundary maps on the same $k$ steps replaced by the background swap (\cref{sec:app-swaps}). 

\paragraph{Expected changes.}
For each case-boundary, we estimate the change under replacement as the negative sum of the attributions over the $k$ replaced steps. The class-map estimate is normalized by the clean class score, while the boundary-map estimate is converted to input steps and divided by $k$. 
In the median over all cases, the class maps sum to 0.97--1.00 of the clean class score.
For boundary attribution, we use the map without boundary-anchored references.

\paragraph{Map agreement.}
Within each setting and replacement side, we rank case-boundaries by the expected class-score change and by the expected boundary displacement.
\Cref{tab:app-map-rank} reports their Spearman correlation with 95\% intervals obtained from 2,000 video-level bootstrap resamples.
For partial correlations, we regress both sets of ranks on the ranks of $k/|I|$, or of
$k/|I|$ and the pyramid level, and correlate the residuals.

The correlations are weak, with no consistent positive association between the two maps across settings and replacement sides, also after controlling for the replaced fraction and pyramid level.

\subsection{Background Swap at the Boundary}
\label{sec:app-swaps}

For each boundary, we replace the $k$ steps immediately inside or outside the predicted interval, where $k=|I|/6$, rounded and clipped to 2--16. 
The background is sampled from another video of the same split, at least 16 steps from any annotation, and results are averaged over three draws.
Outside the interval, the replacement is truncated at the video boundaries; regions with fewer than two remaining steps are excluded.

\paragraph{Effect of the swap.}
Replacing steps inside the interval retracts the boundary by 0.22--0.61 of $k$ and reduces the class score by 7.9--10.2\%.
Outside the interval, the corresponding changes are smaller: 0.05--0.20 of $k$ and 1.9--5.6\%, respectively (\cref{tab:analysis-swap}).

\paragraph{Maps against the swap.}
The correlations in \cref{tab:analysis-swap} have lower 95\% video-bootstrap bounds of at least 0.36 for the class map against score changes and 0.11 for the boundary map against boundary shifts.
Controlling for $k/|I|$ changes the former by at most 0.04, except on EP inside (0.68 to 0.55), where $k/|I|$ itself correlates with the score change at $-0.52$.
We exclude the fused boundary map because the replaced steps coincide with its smallest masked span $H_1^\tau$ for 74\% of inside regions, which could favor agreement by construction.

\paragraph{Shift by class-score change.}
In seven of the eight regions, the 95\% video-bootstrap intervals for the $\leq-10\%$
and $>-2\%$ score-change groups overlap (\cref{tab:app-bins}).
Within the $>-2\%$ group, 63--76\% of case-boundaries exhibit a score increase.
We report inside and outside replacements separately because pooling them could introduce an apparent association driven by their different score-change and boundary-shift distributions.

\end{document}